\documentclass[conference]{ieeeconf}
\usepackage{times}

\usepackage[T1]{fontenc}
\usepackage{multicol}
\usepackage{basicmath}
\usepackage[bookmarks=true]{hyperref}
\usepackage[capitalize]{cleveref}
\usepackage{xcolor}
\usepackage{tabularx}
\usepackage{algorithm}
\usepackage[noend]{algpseudocode}

\usepackage[export]{adjustbox}%
\hypersetup{pdfauthor={Acosta},
	pdftitle={},
	pdfsubject={},
	pdfkeywords={},
	pdfproducer={LaTeX},
	pdfcreator={pdfLaTeX}
}

\begin{document}
	\setlength\arraycolsep{3pt}
	\title{Bipedal Walking on Constrained Footholds with MPC Footstep Control}
	
	\author{\authorblockN{Brian Acosta and Michael Posa}
	\authorblockA{University of Pennsylvania GRASP Lab}}

	\maketitle
	
	\begin{abstract}
		Bipedal robots promise the ability to traverse rough terrain quickly and efficiently, and indeed, humanoid robots can now use strong ankles and careful foot placement to traverse discontinuous terrain. However, more agile underactuated bipeds have small feet and weak ankles, and must constantly adjust their planned footstep position to maintain balance. We introduce a new model-predictive footstep controller which jointly optimizes over the robot's discrete choice of stepping surface, impending footstep position sequence, ankle torque in the sagittal plane, and center of mass trajectory, to track a velocity command. The controller is formulated as a single Mixed Integer Quadratic Program (MIQP) which is solved at 50-200 Hz, depending on terrain complexity. We implement a state of the art real-time elevation mapping and convex terrain decomposition framework to inform the controller of its surroundings in the form on convex polygons representing steppable terrain. We investigate the capabilities and challenges of our approach through hardware experiments on the underactuated biped Cassie. \vspace{-0.2cm}
	\end{abstract}
	
	\IEEEpeerreviewmaketitle
	
	\section{Introduction}\label{sec:intro}
While the ability to traverse unstructured terrain is a key motivation for bipedal robots, navigating these environments is an open problem. Humanoid robots can walk semi-autonomously on discontinuous terrains such as cinder-block piles, but rely on careful foot placement and decoupling footstep planning and balance control \cite{calvertFastAutonomousBipedal2022}, \cite{fallonContinuousHumanoidLocomotion2015}. In contrast, underactuated bipeds have limited ankle torque and small feet, allowing for efficient, agile motion but limiting horizontal center of mass (CoM) actuation. Therefore, footsteps must be continuously re-planned to maintain balance in light of disturbances and model error. 

A simple yet powerful framework for online replanning of stabilizing footstep sequences is to use the step-to-step dynamics of the linear inverted pendulum model (LIP) \cite{kajita3DLinearInverted2001a} to synthesize MPC or LQR footstep controllers. This approach regulates walking speed without ankle torque by using foot placement to affect the initial conditions of each continuous single stance phase. Combined with output tracking via inverse-dynamics based whole body torque controllers, this approach has enabled dynamic and robust walking \cite{xiong3DUnderactuatedBipedal2022}. The Angular Momentum Linear Inverted Pendulum (ALIP) model, in particular, has been shown to accurately describe the bulk motion of walking even for robots with heavy legs \cite{gongOneStepAheadPrediction2021}, and has been used to stabilize walking on sloped terrain \cite{gibsonTerrainAdaptiveALIPBasedBipedal2022}, synthesize specialized stair climbing controllers \cite{dosunmu-ogunbiStairClimbingUsing2023}, and walk on pre-selected constrained footholds \cite{daiBipedalWalkingConstrained2022}. We extend this framework to rough terrain by modeling valid footholds as convex planar polygons, and enforcing via a mixed-integer formulation that each planned footstep lie in a valid foothold. 

\begin{figure}[!t]
	\centering
	\includegraphics[width=0.23\textwidth]{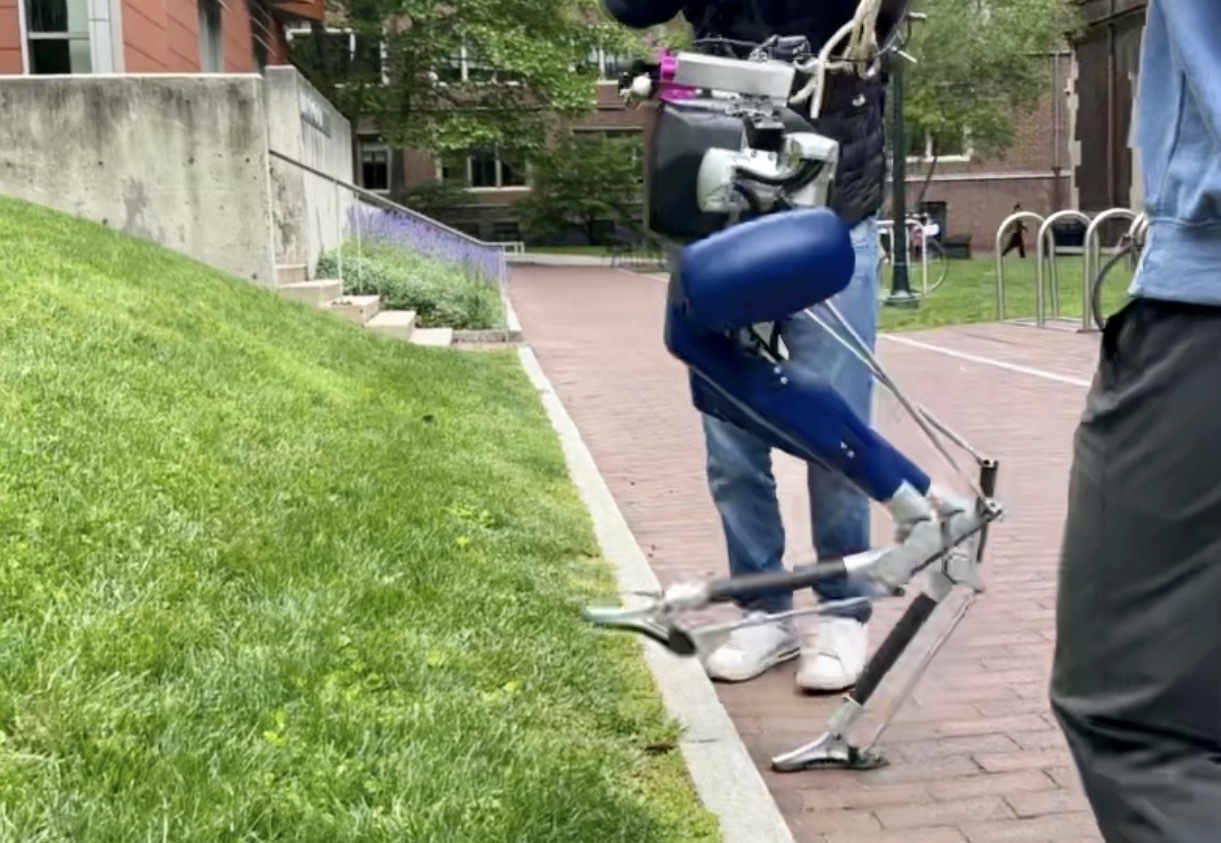}
	\includegraphics[width=0.24\textwidth]{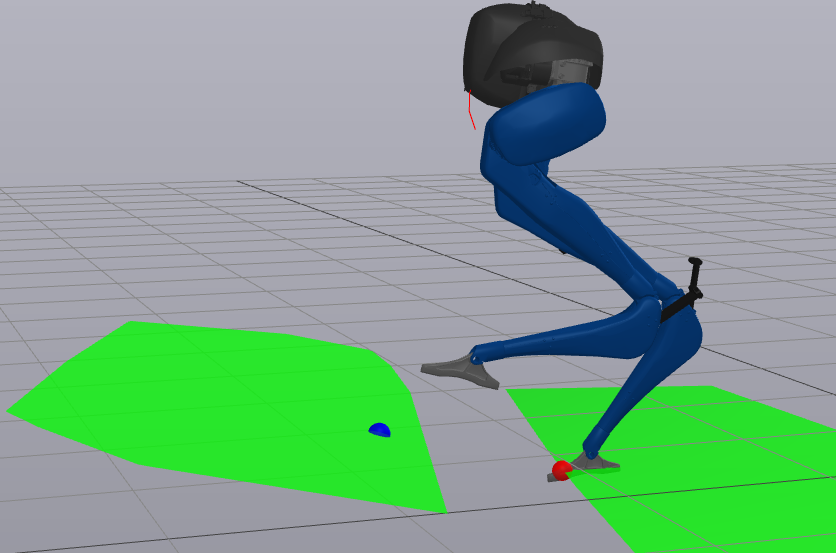}
	\vspace{-0.3cm}
	\caption{We present a model predictive foot placement controller for underactuated bipedal walking with foothold constraints. Left: Cassie stepping over a curb onto a grassy hill using the proposed controller and a real-time terrain segmentation pipeline. Right: visualization of the terrain segmentation and footstep plan during the step-up.\vspace{-0.4cm}}
\end{figure}

We transcribe our controller as a Mixed Integer Quadratic Program (MIQP). MIQPs are used extensively in robot motion planning, but they can be difficult to solve at high rates due combinatorial complexity in the planning horizon. For this reason, existing MIQP footstep planners run at 1-5 Hz \cite{corberesPerceptiveLocomotionWholeBody2023}, limiting their applicability to underactuated walking. In contrast, our controller achieves an average solve time of less than 20 ms even in challenging scenarios by using a low dimensional, linear dynamics model, planning over a short footstep horizon, and heuristically pruning foothold candidates. 

\begin{figure*}[!t]
	\includegraphics[width=0.95\textwidth]{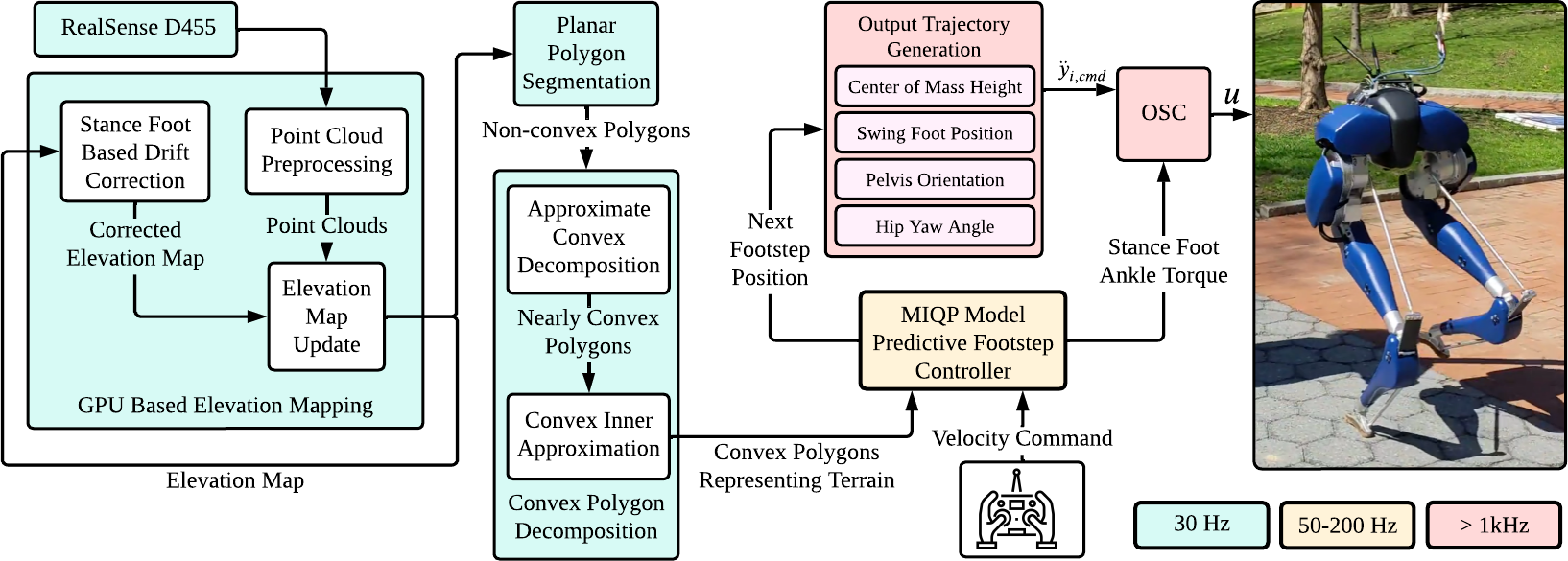}
	\caption{Block diagram of the control stack used to achieve perceptive locomotion on unstructured terrain. A real-time perception pipeline provides convex stepping stone constraints to our proposed MIQP footstep controller, which updates the commanded foot position at 50-200 Hz to maintain balance. This stack allows for foot placement control of underactuated bipeds in previously unseen environments by sensing the terrain on the fly and providing a terrain representation to the walking controller.\vspace{-0.4cm}}
	\label{fig:block_diagram}
\end{figure*}

We use a real-time elevation mapping and terrain decomposition pipeline \cite{mikiElevationMappingLocomotion2022} to represent the terrain as convex polygons online. As the components of this pipeline have been applied primarily to quadrupeds \cite{grandiaPerceptiveLocomotionNonlinear2022} \cite{jeneltenTAMOLSTerrainAwareMotion2022}, we discuss modifications needed for deployment on Cassie, and how certain design choices and properties of the perception stack affect the performance of the MIQP footstep controller. An overview of our perception and control stack can be seen in \cref{fig:block_diagram}. 

The key contributions of this paper are:
\begin{itemize}
	\item An MPC style footstep planner which reasons over discrete foothold selection, footstep sequence, center of mass trajectory, and ankle torque, formulated as a single MIQP which can be solved at up to 200 Hz to stabilize underactuated walking on discontinuous terrain
    \item We extend an existing approach to vision-based real-time elevation mapping and terrain segmentation to be more robust to challenges inherent to underactuated bipedal walking. We introduce a simple algorithm which uses approximate convex decomposition \cite{lienApproximateConvexDecomposition2006} to find a convex polygon decomposition of the steppable terrain. 
	\item Evaluation of the proposed controller on hardware as a full, vision integrated system. We demonstrate perceiving and stepping over curbs in real time, and discuss how perception accuracy limits the robustness of the controller.
\end{itemize}

	\section{Related Work} \label{sec:related}
\subsection{MIQP Footstep Planning}
Deits and Tedrake introduced the use of MIQPs for footstep planning in \cite{deitsFootstepPlanningUneven2014} by decomposing safe terrain into a collection of convex polygons, and using integer variables to assign every footstep to a polygon. Tonneau et al. \cite{tonneauSL1MSparseL1norm2019} provide a convex approximation of this problem as a linear program, and Song et al. \cite{songSolvingFootstepPlanning2021} show how both the mixed integer and linear programming formulations can be made more efficient by using a simplified trajectory planner to prune irrelevant footholds. In contrast to our work, these works only consider geometric and quasistatic stability criteria, and focus on long horizon footstep planning which may take seconds to solve. 

MIQP footstep planning has also been used for quadruped robots. In \cite{risbourgRealtimeFootstepPlanning2022}, Risbourg et al. use the convex relaxation from \cite{tonneauSL1MSparseL1norm2019} online to project the desired footstep sequence to the closest convex footholds, subject to kinematic constraints. In \cite{corberesPerceptiveLocomotionWholeBody2023}, Corberes et al. incorporate this footstep planning strategy as an online foothold scheduler at 1-5 Hz with vision in the loop. Due to the low planning rate, and the lack of dynamics constraints in the contact scheduler, they rely on a separate whole body MPC to find feasible robot trajectories. Aceituno-Cabezas et al. \cite{aceituno-cabezasSimultaneousContactGait2018} formulate a full quadruped trajectory optimization problem using mixed integer constraints for assigning footsteps to footholds and to approximate the nonlinear manifold constraint for 3D rotations. Their trajectory optimization features both kinematic and dynamics constraints, but unlike our work, does not re-plan the footholds in real time.
\subsection{Foot placement control for underactuated walking}
A family of controllers has emerged which use LIP based linear control policies \cite{kimDynamicLocomotionPassiveankle2020} \cite{xiong3DUnderactuatedBipedal2022} or MPC footstep planners \cite{gibsonTerrainAdaptiveALIPBasedBipedal2022}, \cite{xiongGlobalPositionControl2021} to generate footstep plans which are realized by tracking outputs such as CoM height, swing foot position, and joint angles with some form of Quadratic Programming (QP) based whole-body torque control \cite{wensingGenerationDynamicHumanoid2013a} . These controllers view the continuous dynamics of each stance phase as approximated by the autonomous LIP dynamics, and stabilize the walking motion by placing the next footstep in the appropriate position to arrest excess momentum accumulated during single stance. Our controller adopts the same philosophy as these works, but extends the applicability of this approach beyond flat or mildly sloped terrain to terrain which can reasonably be modeled as a collection of convex polygons.
	\section{Preliminaries} \label{sec:prelim}
The controller is implemented in two coordinate frames. $Y$ is the yaw frame, representing a rotation of the world frame about the $z$ axis to match the yaw angle of the floating base. $S$ is the stance frame. It is an identity rotation from the yaw frame, with its origin located at the bottom center of the current stance foot. We follow a $x$-forward, $y$-left, $z$-up convention.

\subsection{Continuous ALIP model}
The ALIP model is an approximation of the CoM dynamics of the robot during single stance based on the LIP \cite{kajita3DLinearInverted2001a}, which uses angular momentum in place of CoM velocity to describe the speed of the robot.  We direct the reader to \cite{gibsonTerrainAdaptiveALIPBasedBipedal2022} for a derivation of the 3D ALIP dynamics assuming piecewise planar terrain with a passive ankle. To take full advantage of Cassie's blade foot, we include ankle torque in the sagittal plane, $u$ as an input to the continuous time ALIP model. The state of the ALIP model consists of the horizontal position of the center of mass and the horizontal components of the angular momentum of the robot about the contact point. The dynamics of the ALIP with ankle torque are given by 
\begin{equation}
\underbrace{\begin{bmatrix} \dot{x}_{com} \\ \dot{y}_{com} \\ \dot{L}_{x} \\ \dot{L}_{y} \end{bmatrix}}_{\dot{x}} =
\underbrace{\begin{bmatrix} 0 & 0 & 0 & \frac{1}{mH}\\
	0 & 0 & \frac{-1}{mH} & 0\\
	0 & -mg & 0 & 0\\
	mg & 0 & 0 & 0 \end{bmatrix}}_{A} \underbrace{ \begin{bmatrix} x_{com} \\ y_{com} \\ L_{x} \\ L_{y} \end{bmatrix}}_{x} + \underbrace{\begin{bmatrix} 0\\0\\0\\1 \end{bmatrix}}_{B} u \label{eq:alip} 
\end{equation} 
where $m$ is the robot's mass, and $H$ is the height of the CoM above the terrain, and all quantities are in the stance frame.
\subsection{ALIP Reset Map}
The reset map enables control of the ALIP through foot placement by relating the positions of the robot's feet to a discrete jump in the ALIP state. On hardware, a double stance phase between footsteps helps avoid oscillations caused by rapidly unloading Cassie's leaf springs. Therefore we derive a reset map from $x_{-}$, the ALIP state just before footfall, to $x_{+}$, the ALIP state just after liftoff, including the double stance phase. We start by integrating the double stance dynamics, and then we apply a coordinate change to express the ALIP state with respect to the new stance foot.

During double stance, we treat the center of pressure (CoP) as a control input, and integrate the resulting dynamics with an assumed input trajectory, 
\begin{align}
p_{CoP} = p_{-} + \frac{t}{T_{ds}}(p_{+} - p_{-})
\label{eq:cop_input}
\end{align}
where $p_{-}$ and $p_{+}$ are the pre- and post-touchdown stance foot positions in the yaw frame, $T_{ds}$ is the duration of double stance, and $t$ is the time since the beginning of double stance. We assume the CoM velocity $v_{CoM}$ is approximately parallel to $p_{+} - p_{-}$. This is reasonable, as the robot is generally stepping in the direction it is walking. Under this assumption, $(p_{CoP} - p_{-}) \times m v_{CoM} \approx 0$, so angular momentum about $p_{-}$ and $p_{CoP}$ are equal.
\begin{equation}
L_{CoP} = L_{p_{-}} +  (p_{CoP} - p_{-}) \times m v_{CoM} \approx L_{p_{-}}.
\end{equation}
By treating the CoP as a virtual contact point and applying \eqref{eq:alip}, we arrive at the continuous dynamics
\begin{equation}
\dot{x} = Ax + \underbrace{
	\begin{bmatrix} 0_{2 \times 1} & 0_{2 \times 1} & 0_{2 \times 1} \\ 0 & mg & 0 \\ -mg & 0 & 0 \end{bmatrix}}_{B_{CoP}}\left(p_{CoP} - p_{-}\right).
\label{eq:reset_cont}
\end{equation}

The solution to \eqref{eq:reset_cont} with the input \eqref{eq:cop_input} and $x(0) = x_{-}$ is a first order hold discretization of \eqref{eq:reset_cont} over double stance, 
\begin{equation}
x(T_{ds}) = A_{r}x_{-} + B_{ds} \left(p_{+} - p_{-}\right).
\label{eq:ds_reset}
\end{equation}
where $A_{r} = \exp(AT_{ds})$ and
\begin{equation}
B_{ds} = A_{r}A^{-1} \left(\frac{1}{T_{ds}} A^{-1} \left(I - A_{r}^{-1}\right)  -A_{r}^{-1}\right)B_{CoP}.
\end{equation} 
The remainder of the reset map is just a coordinate change,
\begin{equation}
x_{+} = x(T_{ds}) + \underbrace{\begin{bmatrix} I_{2\times 2} & 0_{2\times 1} \\  0_{2\times 2} &  0_{2\times 1}\end{bmatrix}}_{B_{fp}} \left(p_{+} - p_{-} \right). \label{eq:alip_reset} 
\end{equation}
with the $fp$ subscript denoting "foot placement".

By sequentially applying \eqref{eq:ds_reset} then \eqref{eq:alip_reset}, we arrive at a reset map from $x_{-}$ to $x_{+}$ which is linear in $x_{-}, x_{+}, p_{-} \text{ and } p_{+}$, 

\begin{equation}
x_{+} = \begin{bmatrix} A_{r} & \left(-B_{ds} - B_{fp}\right) & \smash[b]{\underbrace{\left(B_{ds} + B_{fp}\right)}_{B_{r}}}\end{bmatrix} \begin{bmatrix} x_{-} \\ p_{-} \\ p_{+} \end{bmatrix}.
\label{eq:full_reset}
\end{equation}
	\section{MIQP Model Predictive Footstep Controller}
\begin{figure}
	\centering
	\includegraphics[width=0.25\textwidth]{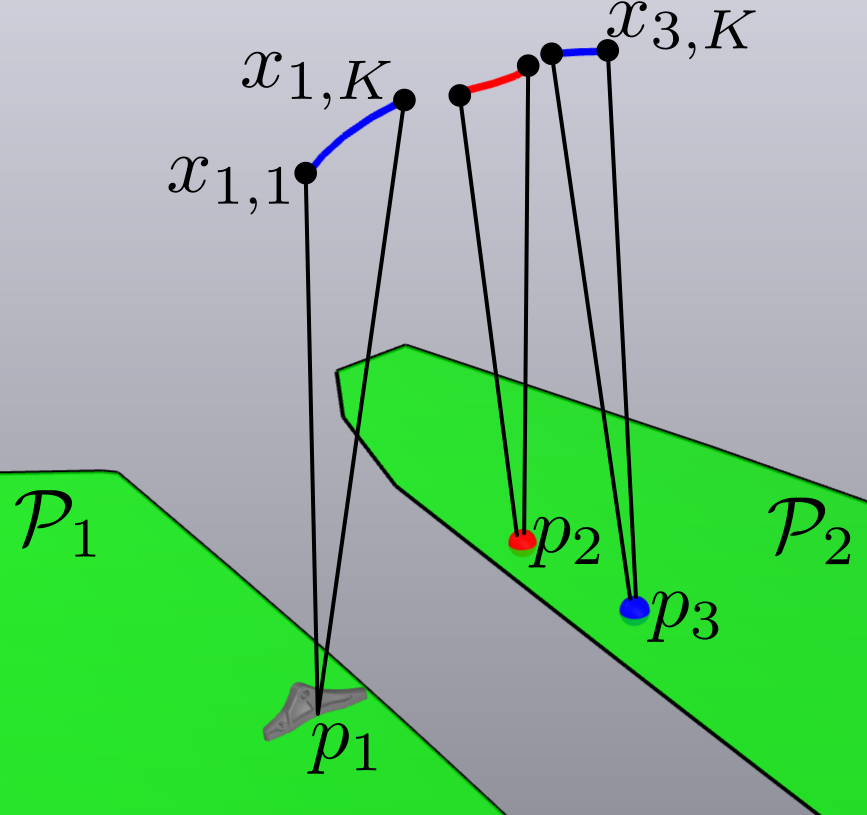}
	\caption{Illustration of key MPFC problem data and decision variables for a horizon of 3 stance phases. The green polygons $\mathcal{P}_{1}$ and $\mathcal{P}_{2}$ are the foothold constraints. The black lines extend from the stance foot to the CoM position at the beginning and end of each single stance phase, and the CoM trajectories for each single stance phase are shown as alternating blue and red paths. The footstep positions are labeled $p_{1}$ -- $p_{3}$, where $p_{1}$ is the current stance position and $p_{2}$ and $p_{3}$ are decision variables.\vspace{-0.4cm}}
	\label{fig:mpc_problem}
\end{figure}

The following section details the formulation of the MPFC as an MIQP. We simultaneously plan the footstep position, input, and ALIP trajectory over a horizon of $N$ stance periods, satisfying the continuous dynamics \eqref{eq:alip} and discrete reset map \eqref{eq:full_reset}. Each footstep is constrained to lie in a steppable region $\mathcal{P}_{i}$, represented as a 2D polygon embedded in 3D space. 
We discretize each single stance period of fixed duration $T_{ss}$ into $K$ knot points in order to apply intra-mode workspace constraints on the center of mass. For convenience, we index state knot points by their stance period and their order within the stance period, so the $k^{th}$ knot point of the $n^{th}$ stance period would be denoted $x_{n,k}$. An illustration of key problem parameters is shown in figure \cref{fig:mpc_problem}.

The MIQP formulation features continuous variables for the state and input trajectories, $\vect{x}$ and $\vect{u}$, as well as the stance foot position for each step, $\vect{p}$. We assign one binary variable, $\mu_{n,i}$ per foothold per step, indicating whether the foothold is used for that step. We can now introduce the problem statement of the MPFC \eqref{eq:mpfc}, and dedicate the rest of this section to elaborating on the cost and constraints. 
\begin{subequations}
\begin{align}
	\begin{split} 
	\underset{\vect{x,u,p, \mu}}{\text{minimize}} \text{  } & \sum_{n = 1}^{N} \sum_{k = 1}^{K-1} \left(\tilde{x}_{n,k}^{T}Q\tilde{x}_{n,k} + u_{n,k}^{T}Ru_{n,k}\right) + \\
	& \qquad \qquad \qquad \qquad \tilde{x}_{N,K}^{T}Q_{f}\tilde{x}_{N,K} 
	\end{split}\label{eq:mpc_cost}\\
	\text{subject to } & x_{n, k+1} = A_{d}x_{n,k} +  B_{d}u_{n,k} \label{eq:mpc_dynamics}\\
	& x_{n+1, 1} = A_{r}x_{n, K} + B_{r}(p_{n+1} - p_{n}) \label{eq:mpc_reset}\\
	& \mu_{n,i} = 1 \implies p_{n} \in \mathcal{P}_{i}  \label{eq:mpc_foothold}\\
	& \sum_{i \in \mathcal{I}} \mu_{n,i} = 1 \label{eq:mpc_sum_int}\\
	& \mu_{n,i} \in \{0,1\}\label{eq:mpc_binary}\\
	& \nonumber \text{CoM, Input, and Footstep limits}
\end{align}

\label{eq:mpfc}
\end{subequations}
where $\vect{\tilde{x}} = \vect{x} - \vect{x_{d}}$ is the error from a desired reference trajectory, and $\mathcal{P}_{i}, i \in \mathcal{I}$ are the footholds. 
\subsection{Dynamics and Reset Map Constraints}
The dynamics constraint \eqref{eq:mpc_dynamics} is the discretization of $\eqref{eq:alip}$ as a sampled system. Letting $\Delta t = T_{ss} / (K-1)$, then
\begin{align}
A_{d} = \exp(A \Delta t)\\
B_{d} = A^{-1}\left(A_{d} - I\right)B.
\end{align}
The reset map \eqref{eq:mpc_reset} is \eqref{eq:full_reset} expressed with MPFC indexing.
\subsection{Foothold Constraints}
Each convex polygonal foothold is defined by a plane $f_{i}^{T}p = b_{i}$ and a set of linear constraints $F_{i}p \leq c_{i}$. The logical constraint \eqref{eq:mpc_foothold} is enforced with the big-M formulation
\begin{subequations}
\begin{align}
	F_{i}p_{n} \leq c_{i} + M(1 - \mu_{n,i})\\
	f_{i}^{T}p_{n} \leq b_{i} + M(1 - \mu_{n,i})\\
	-f_{i}^{T}p_{n} \leq -b_{i}+ M(1 - \mu_{n,i}).
\end{align}
\label{eq:bigM}
\end{subequations}
With appropriately normalized $F_{i}$ and $f_{i}$, \eqref{eq:bigM} corresponds to relaxing each foothold constraint by $M$ meters when $\mu_{i} = 0$. Since our problem scale is on the order of 2 m, we choose M = 10 for simplicity\footnote{$M$ must be large enough for every relaxed foothold to contain every unrelaxed foothold, but should otherwise be small for numerical stability}. The binary constraint \eqref{eq:mpc_binary} and the summation constraint \eqref{eq:mpc_sum_int} imply that exactly one foothold must be chosen, and the remaining footholds relaxed. 
\subsection{State, Input, and Footstep Limits}
We add a bounding box constraint on the CoM position with conservative bounds on $y_{CoM}$ to avoid hip-roll joint limits. We limit the ankle torque to 5 Nm, which is what the OSC can reliably provide without foot slip. Finally, we add a constraint to prevent the feet from crossing the $x-z$ plane.

\subsection{Reference Design}
Given a desired average horizontal velocity, $v_{d} \in \mathbb{R}^{2}$, we generate a reference trajectory for the MPFC by finding a periodic ALIP trajectory which achieves this average velocity with a user-specified stance width $l$.  In addition to letting us tune the stance width directly, this also ensures the desired position and angular momentum are consistent with the ALIP dynamics without using ankle torque. First, the gait parameters $v_{d}$ and $l$ are encoded into a footstep sequence 
\begin{equation}
p_{n+1} = p_{n} + v_{d}T_{s2s} + \sigma_{n} l \hat{e}_{y} \label{eq:vdes}
\end{equation} where $T_{s2s} = T_{ss} + T_{ds}$ and $\sigma_{n} =  -1, 1$ for left and right stance, respectively. To find the corresponding periodic ALIP trajectory, we define the step-to-step dynamics, which combine the single stance dynamics and reset map to arrive at a discrete dynamical system which has state $x_{n,1}$, the ALIP state at the beginning of single stance, and takes ${}^Sp_{n+1}$ as an input. The dynamics are
\begin{equation}
x_{n+1, 1} = \exp(A T_{s2s})x_{n, 1} + B_{r}{}^Sp_{n+1}.
\label{eq:discrete}
\end{equation}
We find the reference gait by substituting \eqref{eq:vdes} into \eqref{eq:discrete} and rolling out the dynamics for two stance modes, then solving for $x_{1,1} = x_{3,1}$ \footnote{This is a specific choice of Period-2 orbit \cite{xiong3DUnderactuatedBipedal2022} which achieves symmetry between left and right stance.}.

	\section{Output Tracking via Operational Space Control} \label{sec:osc}
To realize the planned walking motion on the physical robot, MPC outputs are tracked with an inverse-dynamics based operational space controller (OSC). We use the same quadratic program as our previous Cassie examples \cite{yangImpactInvariantControl2021} \cite{chenOptimalReducedorderModeling2020b}. This section describes the construction of the task space trajectories tracked by the OSC.

\subsection{Center of Mass Reference}
Given a footstep plan from the MPC, we construct a CoM trajectory which enforces the local planarity assumption of the ALIP model by constructing the least-inclined plane passing through the current and imminent stance foot positions. Letting $p = {}^S p_{n+1}$, the plane parameters are the solution to

\begin{equation}
	\begin{bmatrix}
		p_{x} & p_{y}\\ -p_{y} & p_{x}
	\end{bmatrix} 
	\begin{bmatrix}
	k_{x} \\ k_{y}
	\end{bmatrix} = 
	\begin{bmatrix}
	p_{z} \\ 0 
	\end{bmatrix}.
\end{equation}
After solving for $k_{x}$ and $k_{y}$, we define the reference trajectory for the CoM height in the stance frame as
\begin{equation}
z_{c}(t) = H + k_{x} x_{c}(t) + k_{y} y_{c}(t).
\end{equation}

\subsection{Swing Foot Reference}
\begin{figure}
	\centering
	\includegraphics[width=0.25\textwidth]{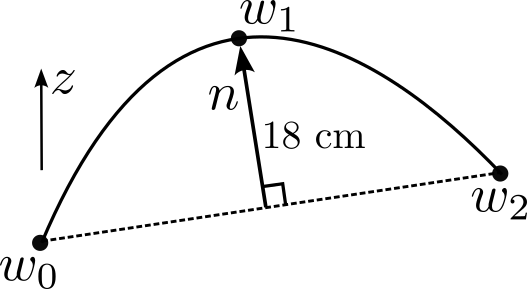}
	\caption{We realize the planned footstep footstep positions by constructing a swing foot trajectory through the way points $w_{0}$, $w_{1}$, and $w_{2}$, where $w_{0}$ is the position of the swing foot at the beginning of the single stance phase, and $w_{2}$ is the most recent MPFC solution for the incoming stance foot. \vspace{-0.4cm}}
	\label{fig:swing_traj}
\end{figure}

We generate swing foot trajectories by constructing a spline between the initial and final foot location during swing. First we generate an additional way point above the line connecting the initial and final foot location, as shown in \cref{fig:swing_traj}. Then,
as a heuristic for generating swing foot trajectories which will be tractable to track despite Cassie's leaf springs, we construct a minimum-snap spline through the way points. The motivation for this is that when considering the spring dynamics, the foot position is relative degree four to the motor torques. 

\subsection{Constant references} 
We track a constant pelvis roll and pitch of zero, and a constant hip yaw (abduction) angle of zero. We track a commanded pelvis yaw rate from the remote control, and a swing toe angle so that Cassie's foot makes an angle of $\arctan{k_{x}}$ with the ground.  

	\section{Perception Stack for Hardware Experiments}
Here we describe the perception stack used to translate point clouds from the Intel RealSense D455 depth camera to convex foothold constraints in real time. The perception stack architecture is shown in \cref{fig:block_diagram}. 

\subsection{State Estimator}
We use the contact-aided invariant extended Kalman filter developed by Hartley et al. \cite{hartleyContactaidedInvariantExtended2020} to estimate the pose and velocity of the floating base. Due to the unobservability of Cassie's global position, we experience state estimator drift, especially vertically, which is accounted for in the elevation mapping node as described below.

\subsection{RealSense D455 Depth Camera}
\begin{figure*}[t]
	\centering
	\includegraphics[width=0.19\textwidth]{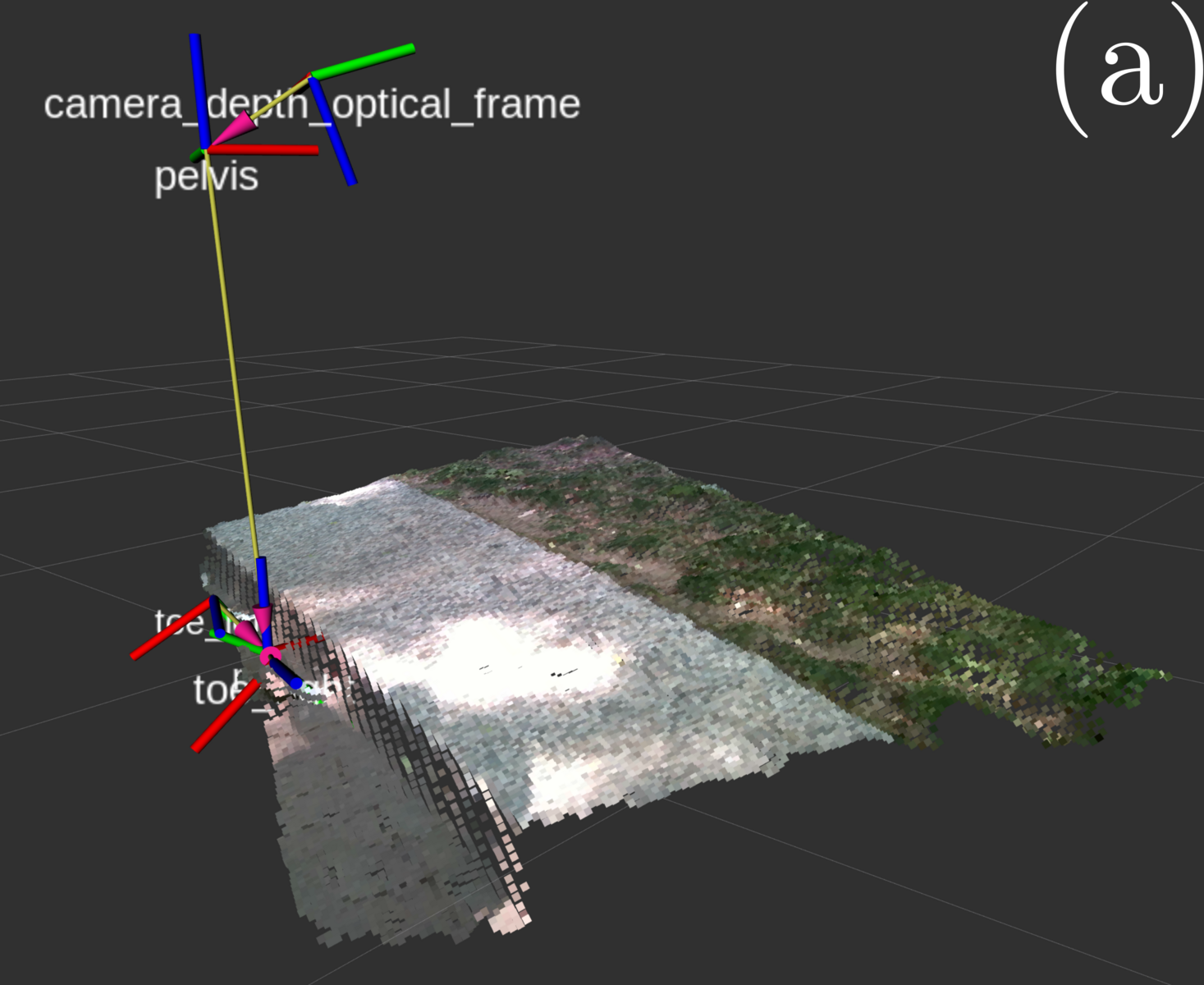}
	\includegraphics[width=0.19\textwidth]{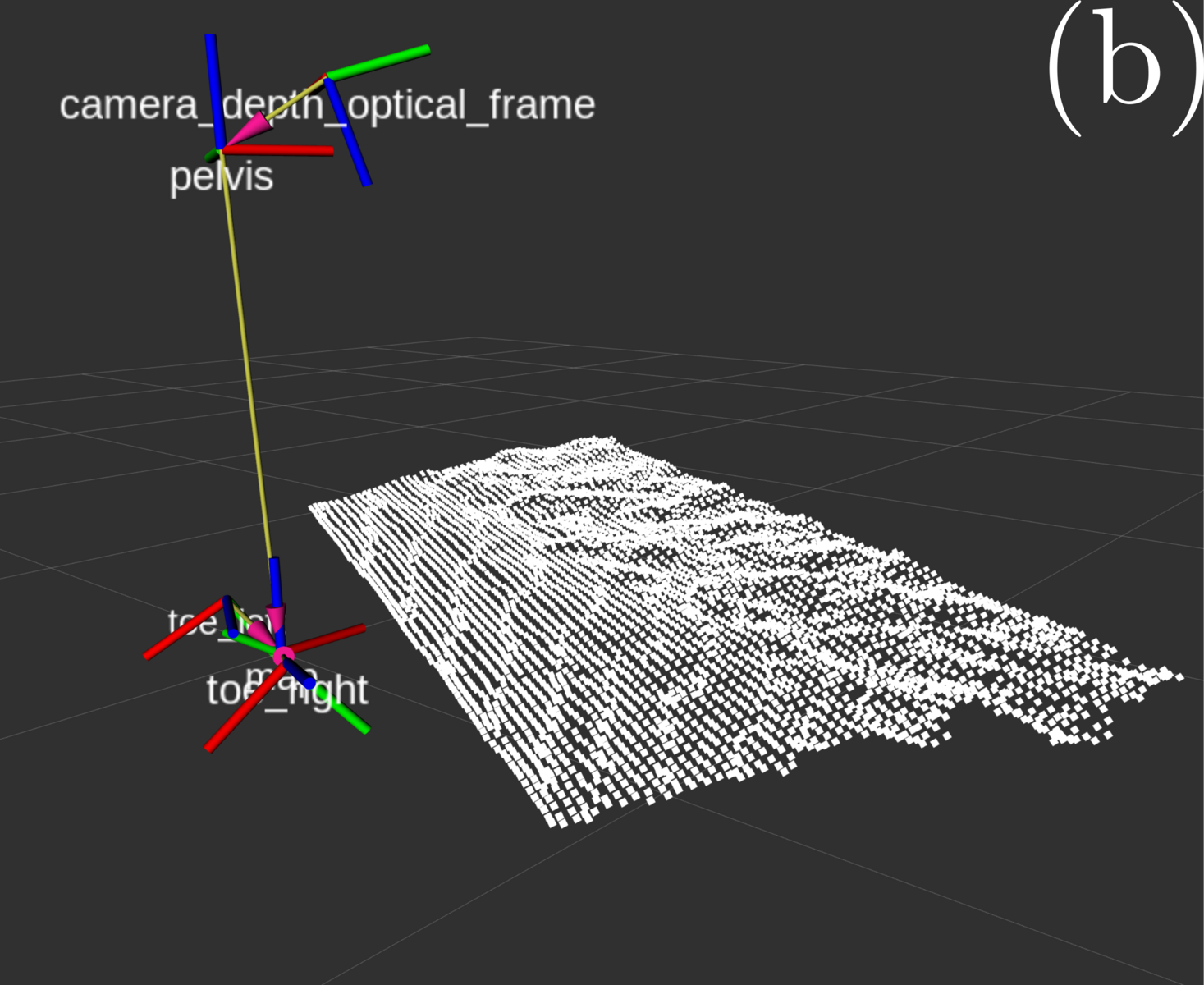}
	\includegraphics[width=0.19\textwidth]{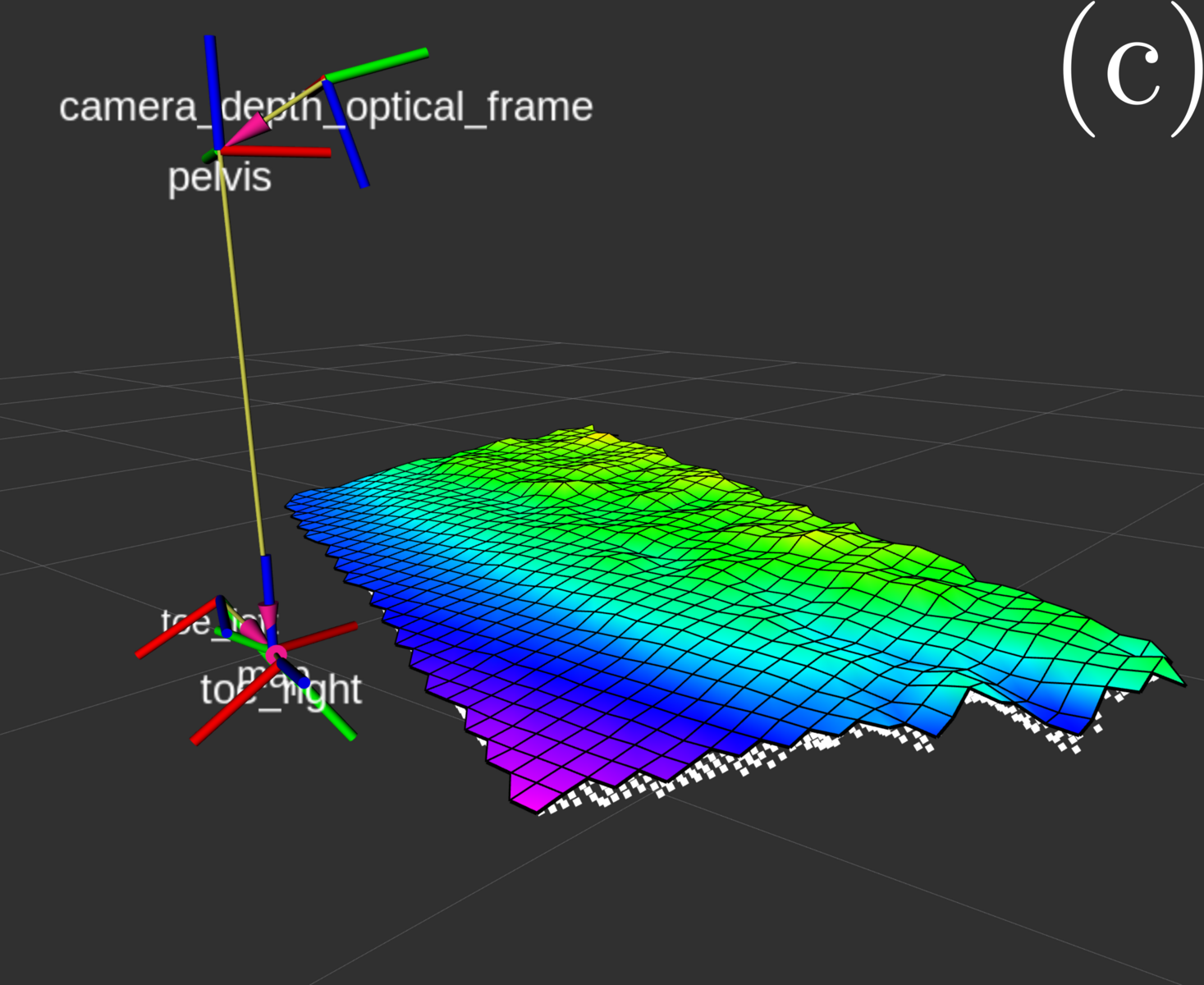}
	\includegraphics[width=0.19\textwidth]{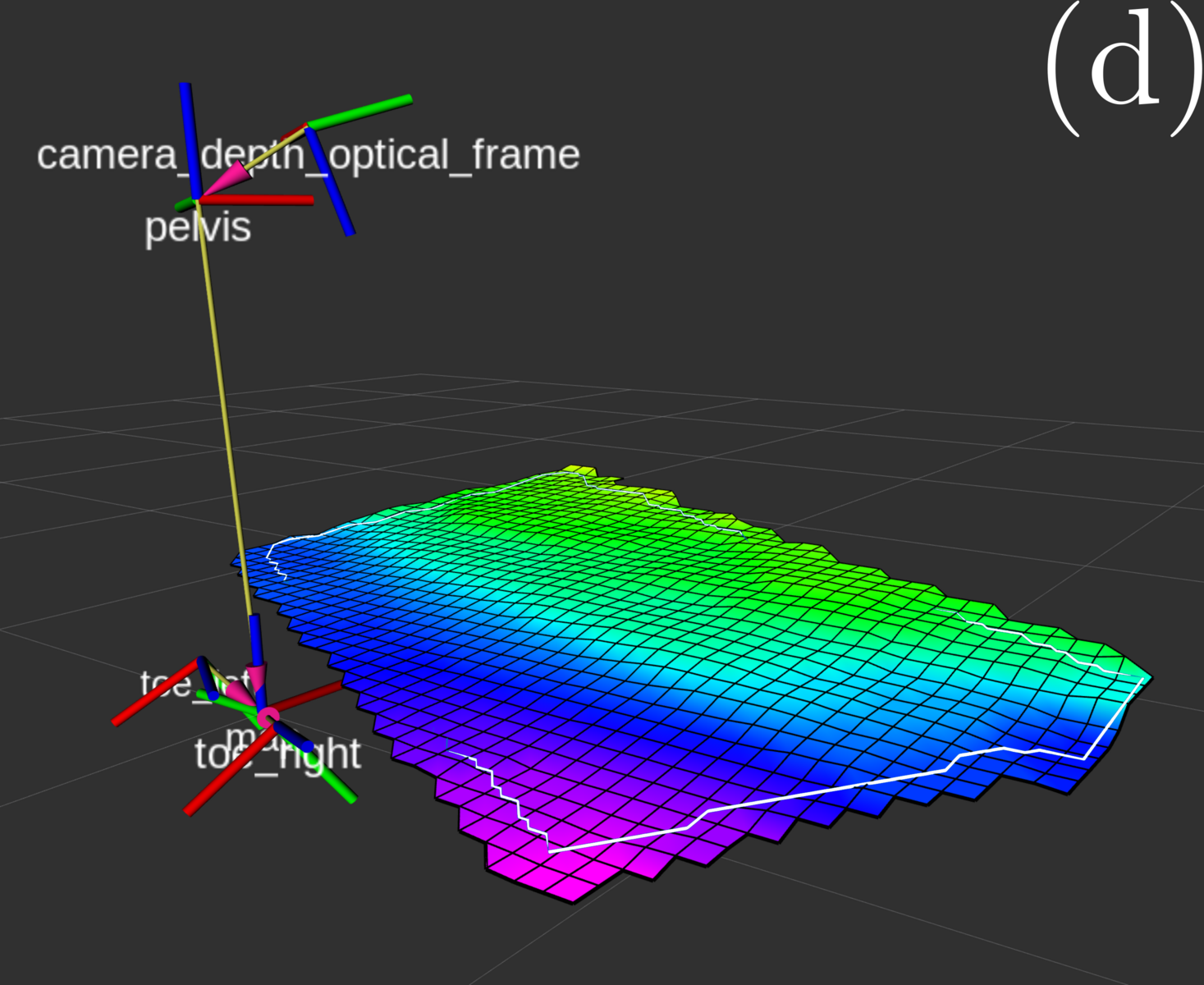}
	\includegraphics[width=0.19\textwidth]{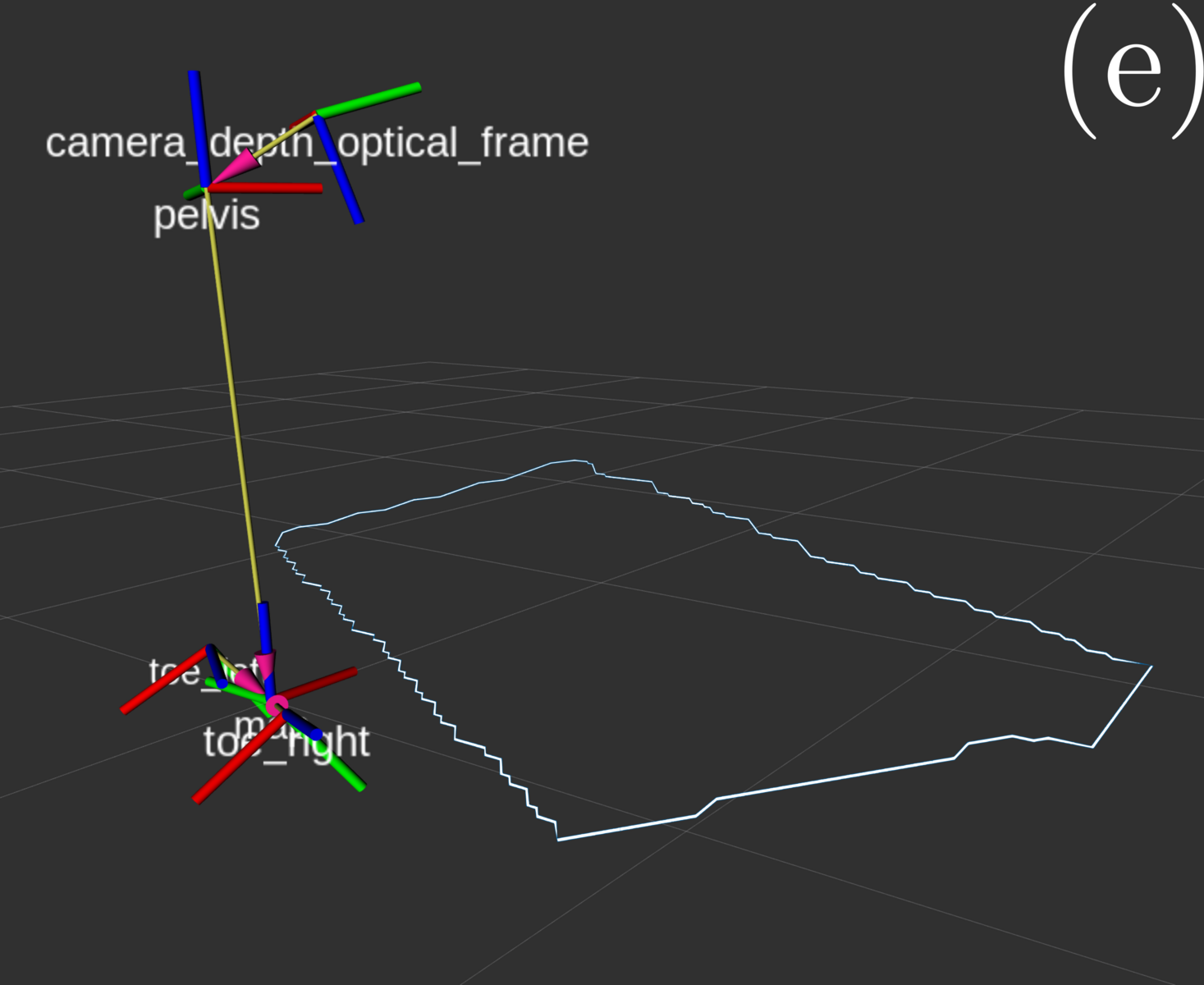}
	
	\caption{Example of the elevation mapping and planar polygon extraction process showing the field of view of the Intel RealSense. Frame (a) shows the raw point-cloud data as seen in the world frame. Our point cloud pre-processing step (b) crops out potentially noisy measurements of Cassie's front shell and a bounding box around the robot's feet. Frame (c) shows the raw elevation map constructed from the cropped point cloud. Frame (d) Shows the map after applying a Gaussian blur and median filter, as well as the steppable planar polygon extracted from the map. (e) Shows the extracted polygon by itself.}
	\label{fig:pointcloud_pipeline}
\end{figure*}

The RealSense is mounted to Cassie's pelvis, looking down at the terrain in front of the robot (\cref{fig:pointcloud_pipeline}). We use the \texttt{realsense-ros}\footnote{https://github.com/IntelRealSense/realsense-ros} ROS package to publish point-cloud data at 30 Hz, applying a decimation filter to reduce the number of points sent to the elevation mapping node.

\subsection{GPU Based Elevation Mapping}
We use the GPU based elevation mapping framework developed by Miki et al. in \cite{mikiElevationMappingLocomotion2022} to construct a robot-centric elevation map of the terrain. This framework represents the terrain as a grid around the robot, with the height of each cell updated by point cloud measurements through a Kalman filter. The quality of the convex planar decomposition, and ultimately the stability of the controller, depends on the accuracy of the elevation map, so we make several Cassie-specific modifications to \cite{mikiElevationMappingLocomotion2022}, outlined below.

\subsubsection{Point Cloud Preprocessing}
To eliminate spurious measurements of the Cassie's front shell, we crop out a band of points along the near edge of the depth camera frame. Additionally, we crop out points outside user-specified minimum and maxium depths. We mask out Cassie's legs by removing all points from a bounding box extending up and back from the front of each foot. 

\subsubsection{Drift Correction using the Stance Foot}
While \cite{mikiElevationMappingLocomotion2022} features a floating base drift correction feature which compares the height of the input point cloud to the height of the existing map, we found this to be insufficient for the severity of floating base drift we experience on Cassie. Because Cassie's state estimate experiences a consistent upward drift due to impacts during touchdown, and because we have only one depth camera on the front of the robot, terrain under and behind the robot is estimated to be lower than in reality. To correct for this, before each point cloud update, we adjust the height of the elevation map by adding the height difference between the elevation map and the current stance foot. 

\subsection{Planar Segmentation}
We use the planar segmentation module provided by \cite{mikiElevationMappingLocomotion2022} (but described in \cite{grandiaPerceptiveLocomotionNonlinear2022}) to segment the elevation map into planar polygons. First, several filters are applied to the elevation map. In addition to the de-noising median filter described in \cite{grandiaPerceptiveLocomotionNonlinear2022}, we apply an erosion filter and Gaussian blur to the height map to smooth out the terrain into its broad features. Next, each pixel in the elevation map is classified as steppable or not based on surface inclination and roughness in a neighborhood around the pixel. From this classification, connected components of steppable terrain are identified, and their outline is extracted as a 2D polygon embedded in 3D space, with a safety margin of 5 cm. It should be noted that the effective safety margin is higher, as the Gaussian blur rounds off sharp corners in the elevation map.

\begin{figure}[t]
	\centering
	\includegraphics[width=0.15\textwidth]{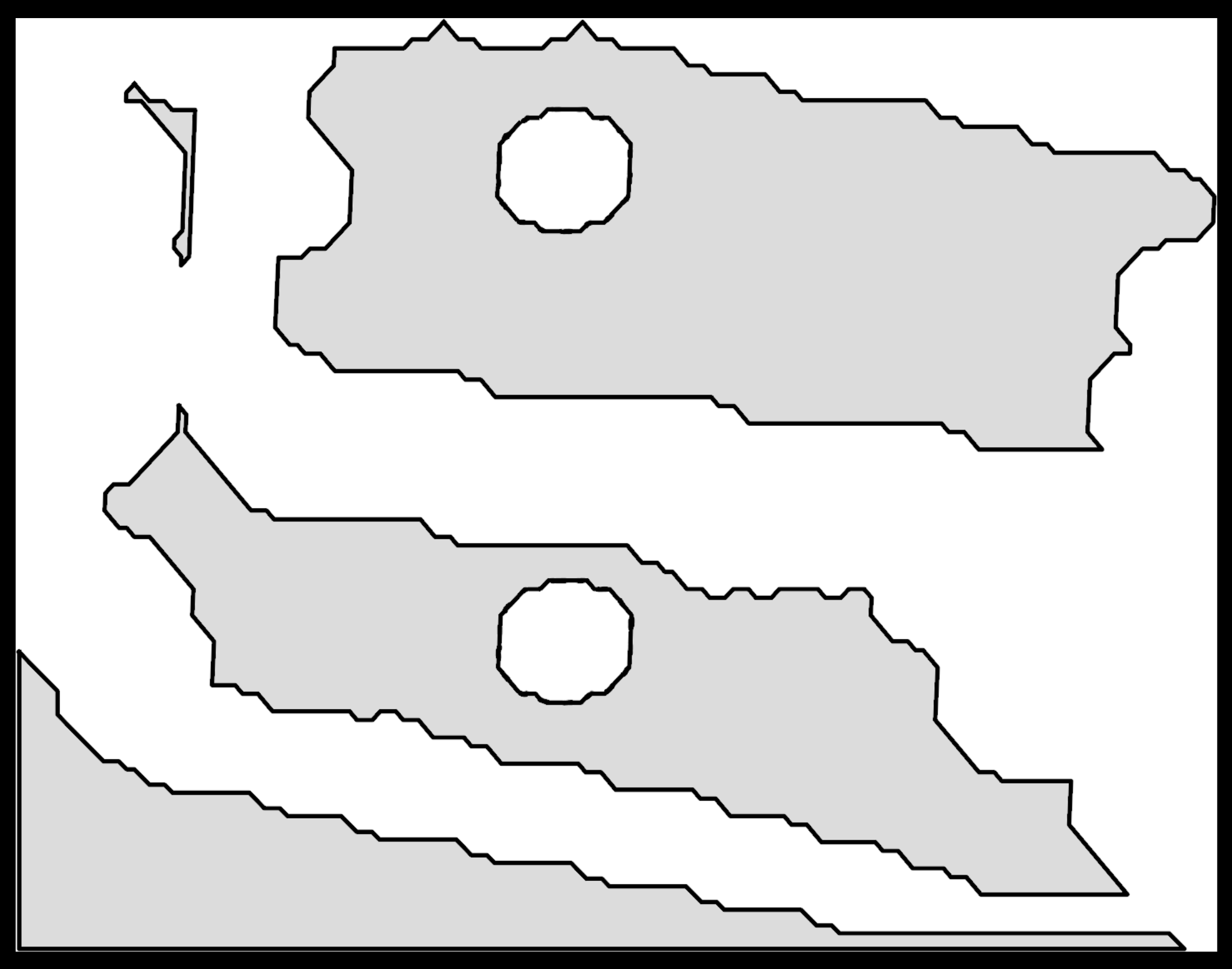}
	\includegraphics[width=0.15\textwidth]{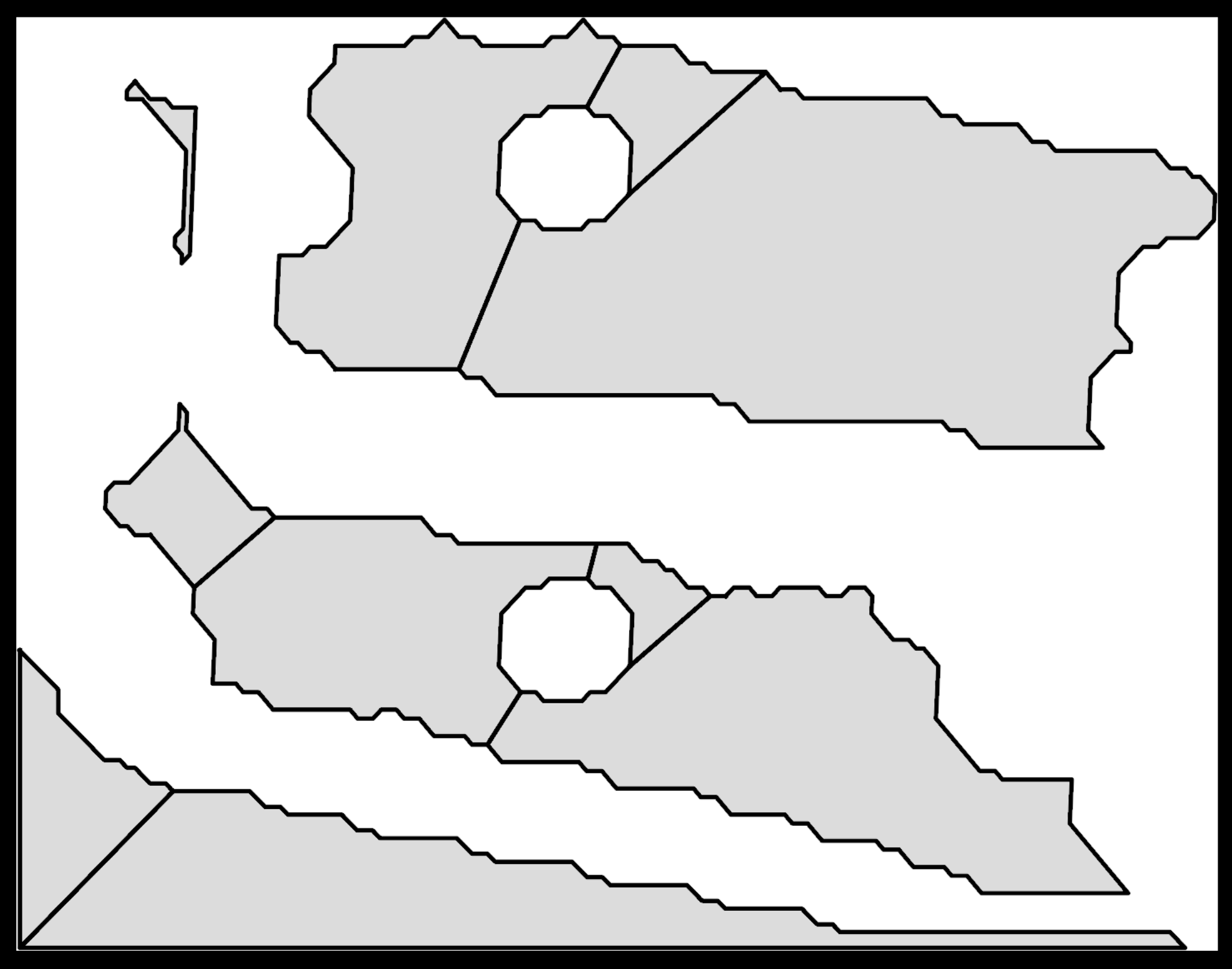}
	\includegraphics[width=0.15\textwidth]{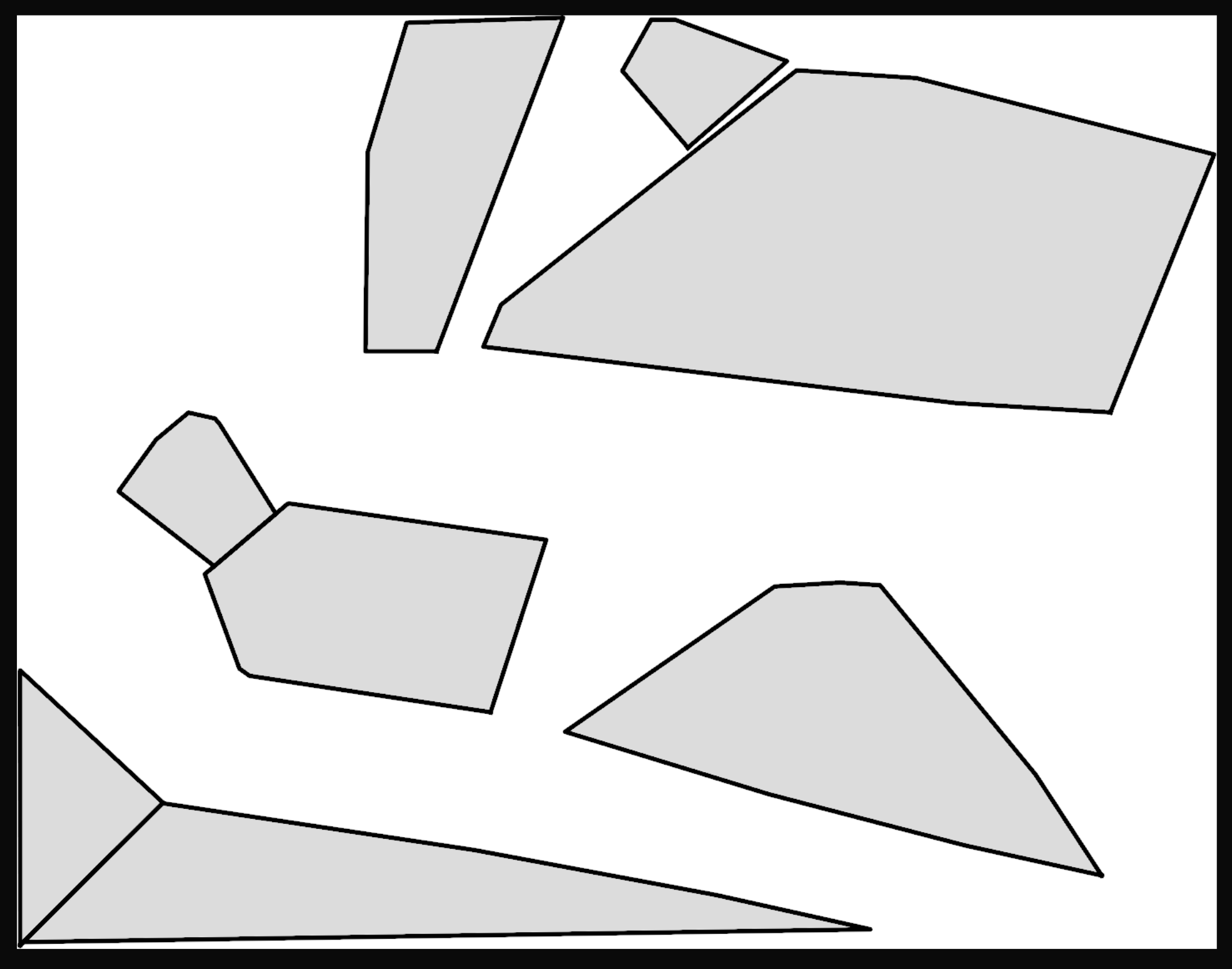}
	\caption{Illustration of the convex polygon decomposition process. Left: original nonconvex polygons with holes. Middle: approximate convex decomposition of the original polygons. Right: convex inner approximation of the approximately convex components, with small components filtered out. When tuning the convex decomposition, there is a trade off between maximizing the total traversable area and maintaining a low number of convex polygons.\vspace{-0.4cm}}
	\label{fig:convex_decomp}
\end{figure}

\subsection{Convex Polygon Decomposition}
In general, the planar segmentation yields non-convex polygons with holes (caused, for example, by small obstacles or other unsteppable areas), but we require convex foothold constraints for the MPFC. We use a two stage process (\cref{fig:convex_decomp}) to find a set of convex polygons whose union approximately matches the original non-convex polygon. This avoids creating many small triangles like an exact convex decomposition would, decreasing the number of integer variables in the MPFC.

First, we perform approximate convex decomposition (ACD)\cite{lienApproximateConvexDecomposition2006} on each polygon. ACD returns a decomposition of the original region into polygons which are $\tau$-approximately convex, with $\tau$ representing the depth of the largest concave feature. After filtering out polygons with area less than 0.1 $\text{m}^{2}$, we find a convex inner-approximation of these nearly convex polygons with a greedy approach we name the whittling algorithm (\cref{alg:whittling}), after the way it makes incremental cuts to the polygon. We initialize the output polygon, $\mathcal{P}$ as the convex hull of the original polygon, then take the intersection of $\mathcal{P}$ with greedily chosen half-spaces until no vertices of the original polygon are contained in $\mathcal{P}$. While this does not guarantee containment of $\mathcal{P}$ in the original polygon, we do not see violations in practice.
\begin{algorithm}[H]
	\caption{Whittling Algorithm} \label{alg:whittling}
	\begin{algorithmic}[0]
		\Require Input polygon vertices $V = \{v_{0}\ldots v_{n}\}$
		\Procedure{Whittle}{$V$}
		\State $\mathcal{P} \gets \text{ConvexHull}(V)$
		\ForAll{$v_{i}$}
		\If{$v_{i} \in \text{Interior}(\mathcal{P})$}
		\State $H$ = MakeCut($v_{i}$, $\mathcal{P}$)
		\State $\mathcal{P} \gets \mathcal{P} \cap H$
		\EndIf
		\EndFor
		\State \Return $\mathcal{P}$
		\EndProcedure
	\end{algorithmic}
\end{algorithm}

\(\text{MakeCut}(\mathcal{P}, v)\) is a QP which finds $a$ such that the half-space $H  = \{x \mid a^{T}(x - v) \leq 0\}$ contains as much of $\mathcal{P}$ as possible, as measured by minimizing the squared hinge loss $\sum \max(a^{T}(p_{i} - v), 0)^{2}$, where $p_{i}$ are the vertices of $\mathcal{P}$. 

	\section{Results}
Both layers of the control stack are implemented in C++ using the Drake \cite{russtedrakeandthedrakedevelopmentteamDrakeModelBasedDesign2019} systems framework and mutlibody kinematics/dynamics. The MPFC and OSC use Drake's interfaces to the Gurobi and OSQP solvers respectively. The controllers are run in separate processes and communicate over LCM \cite{huangLCMLightweightCommunications2010}.
This abstracts the controller from source of the robot state information, allowing us to test identical code in simulation or on hardware. We use a horizon of 3 stance periods for the MPFC in order to plan multiple footsteps ahead.

\subsection{Simulation Experiments}
\begin{figure}[h]
	\centering
	\includegraphics[width=0.23\textwidth]{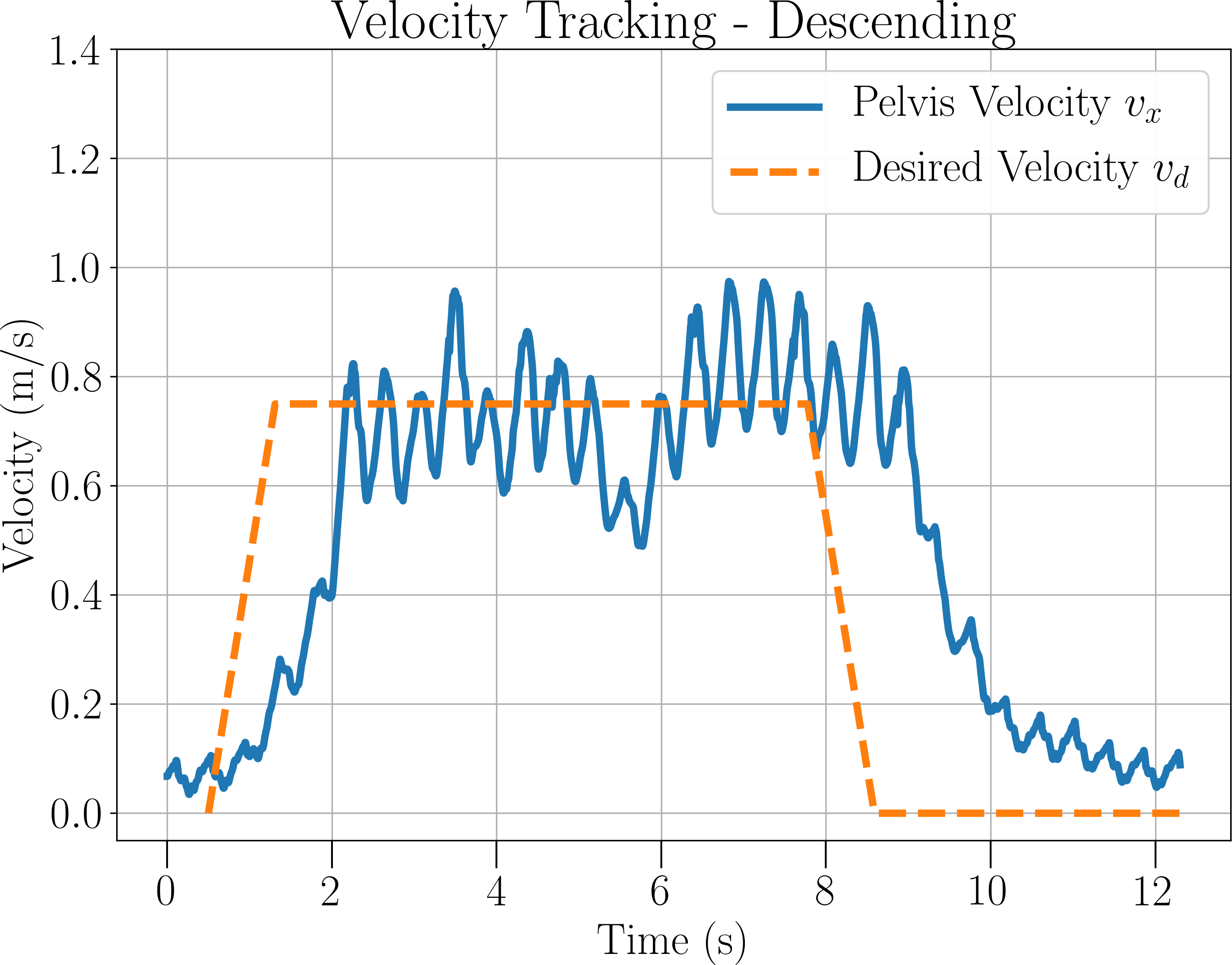}
	\includegraphics[width=0.23\textwidth]{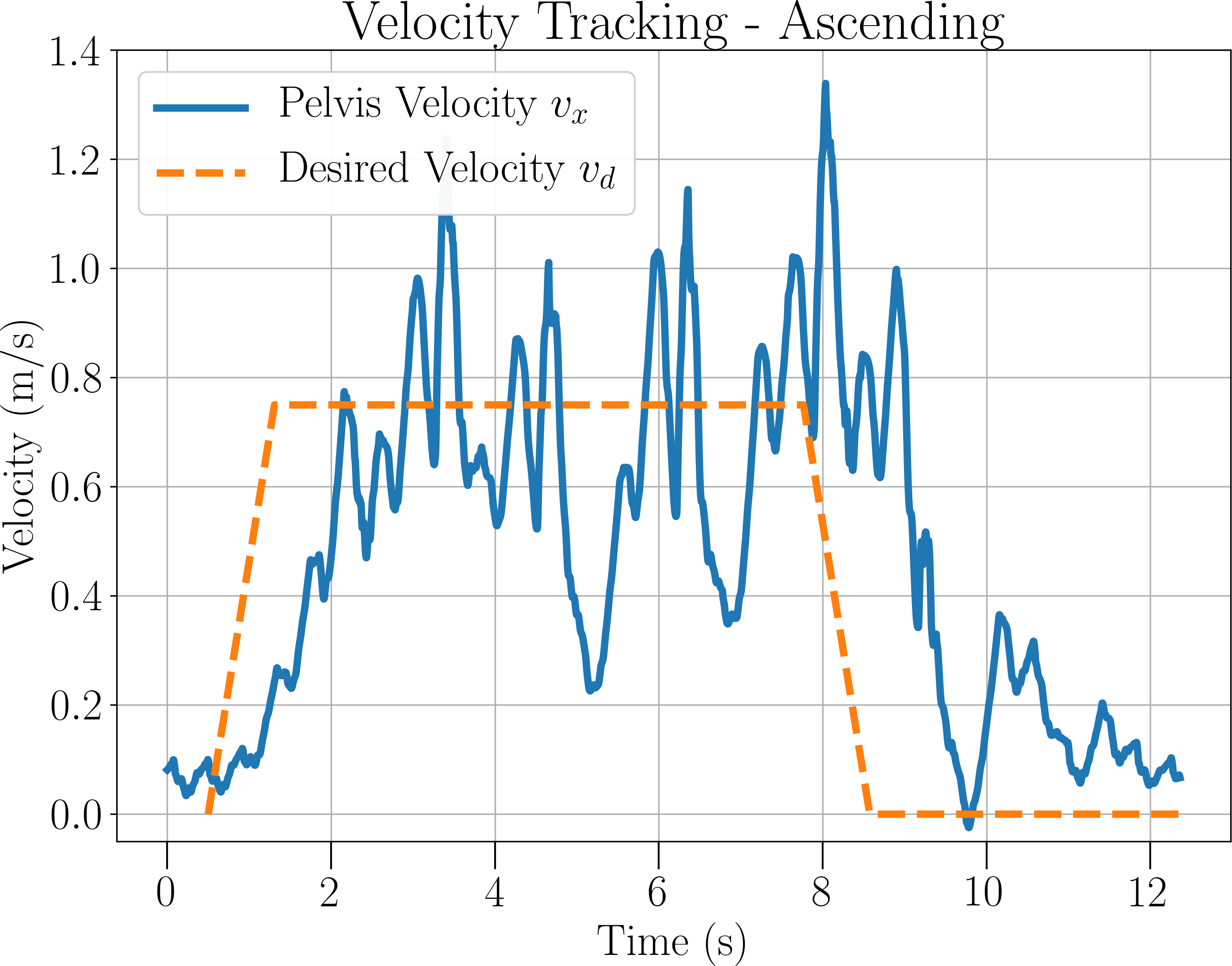}
	\includegraphics[width=0.23\textwidth]{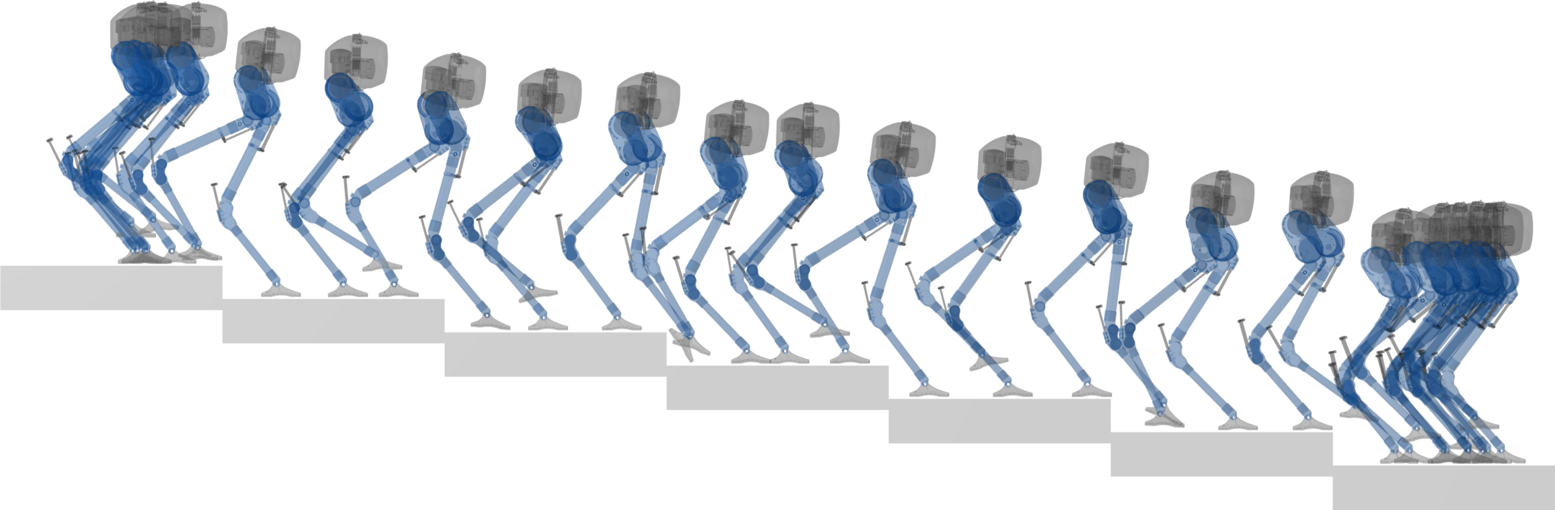}
	\includegraphics[width=0.23\textwidth]{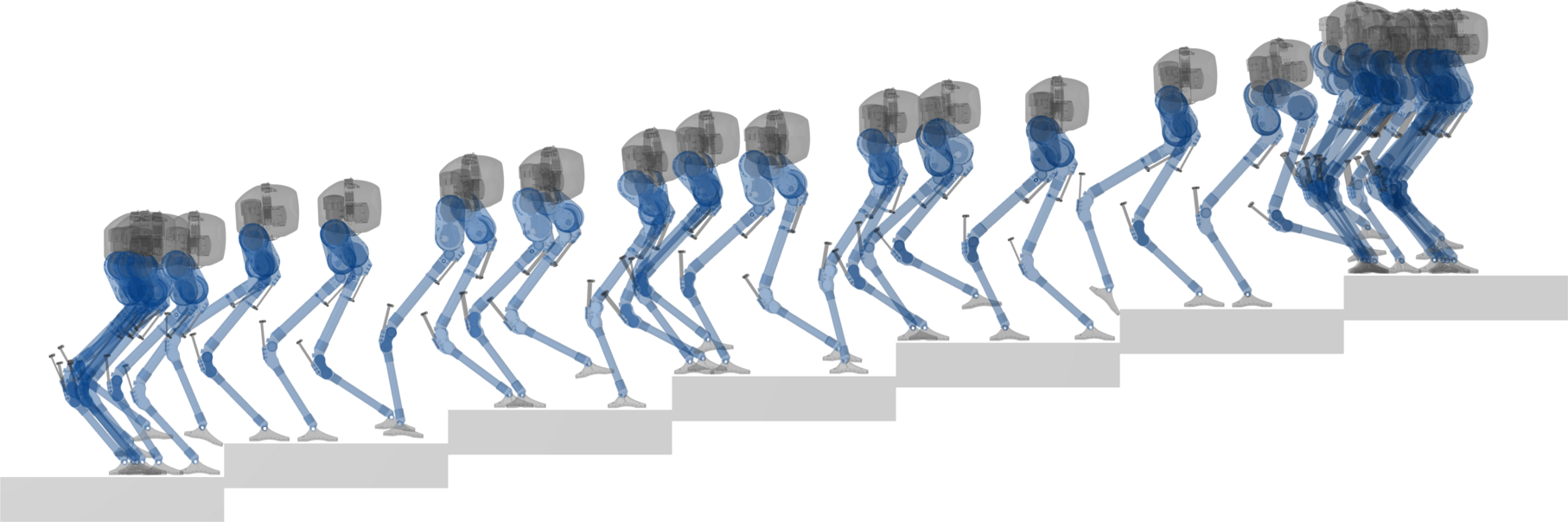}
	\caption{Cassie descending and ascending a set of 1 m. long, 15 cm. tall steps with a pre-programmed velocity profile in simulation. Cassie tracks the commanded velocity with occasional deviations to satisfy foothold constraints.\vspace{-0.2cm}}
	\label{fig:sim_stairs}
\end{figure}

To highlight the capabilities of the controller in an idealized environment, we demonstrate Cassie ascending and descending steps in simulation (\cref{fig:sim_stairs}). While we model Cassie as realistically as possible, including springs, reflected inertia, motor curves, and realistic joint limits, the MPFC and OSC have access to ideal state and terrain information. Cassie is able to track a commanded velocity of 0.75 m/s on meter long stairs, and automatically deviate from the velocity command to satisfy foothold constraints. Given Cassie's limited ankle torque, walking can only be stabilized on steps long enough to use footstep placement as the primary control input, 1 m for a commanded speed of  0.75 m/s, and 50 cm for a commanded speed of 0.5 m/s. Additionally, for steps taller than 15 cm, the ALIP model becomes a poor approximation of the dynamics, and the trailing swing foot starts to clip the stairs when stepping up. We witnessed this effect more severely on hardware, and discuss it further in \cref{sec:challenge}.

\subsection{Hardware Experiments}
For hardware deployment, we run the high speed state estimator and OSC loops on Cassie's onboard NUC computer and send joint torque commands to Cassie's target PC over UDP. The NUC is upgraded from stock to achieve better performance \footnote{https://github.com/DAIRLab/cassie\_documentation/wiki/Upgrading-the-Intel-NUC}. The MPFC and perception stack run on an offboard ThinkPad p15 Laptop with an 8-core, 2.3 GHz Intel 1180H processor and 24 GB of RAM. The laptop is carried in a backpack by one of the safety bar carriers and networked with the NUC over ethernet for LCM and ROS communication. 

\begin{figure*}[t]
	\centering
	\includegraphics[width=0.1\textwidth]{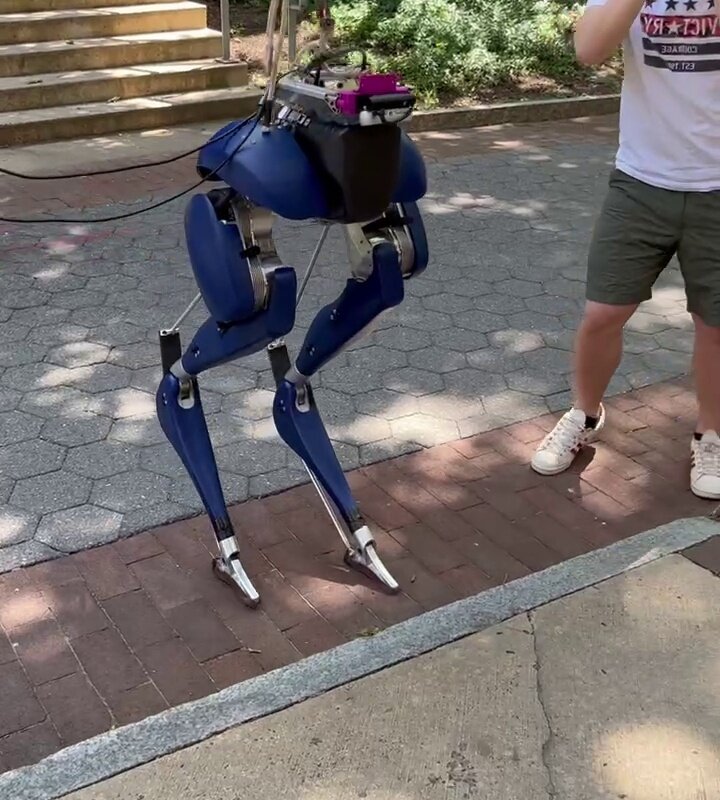}
	\includegraphics[width=0.1\textwidth]{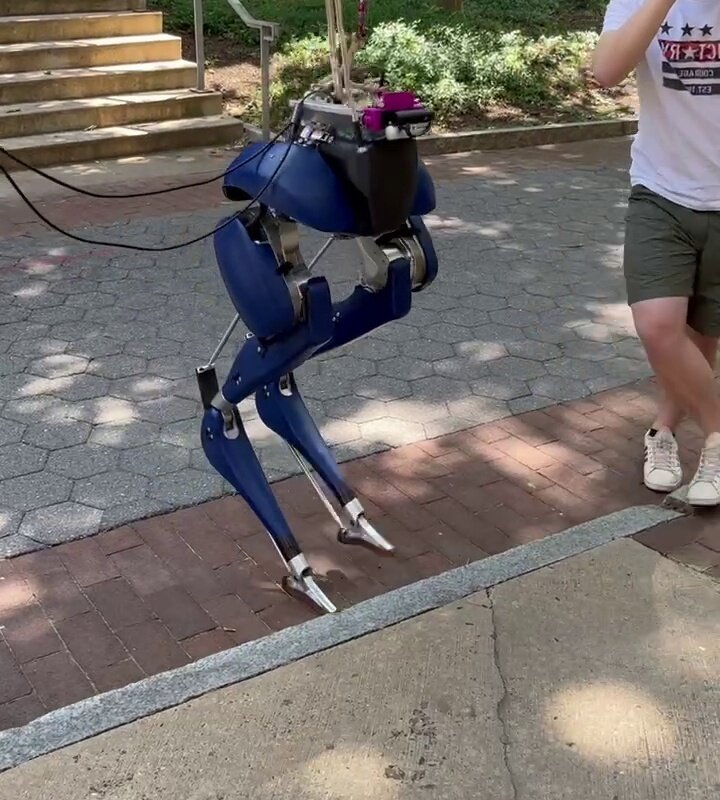}
	\includegraphics[width=0.1\textwidth]{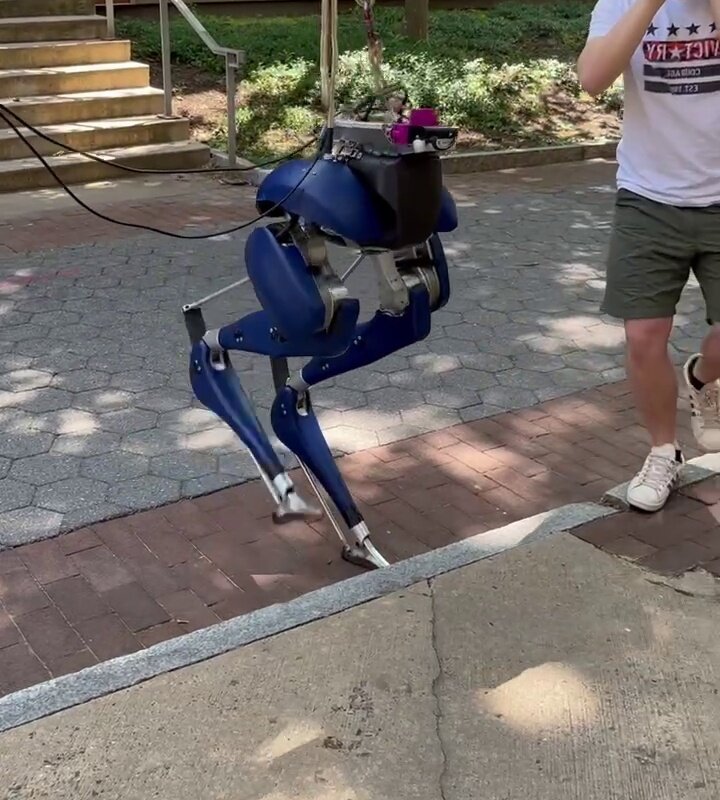}
	\includegraphics[width=0.1\textwidth]{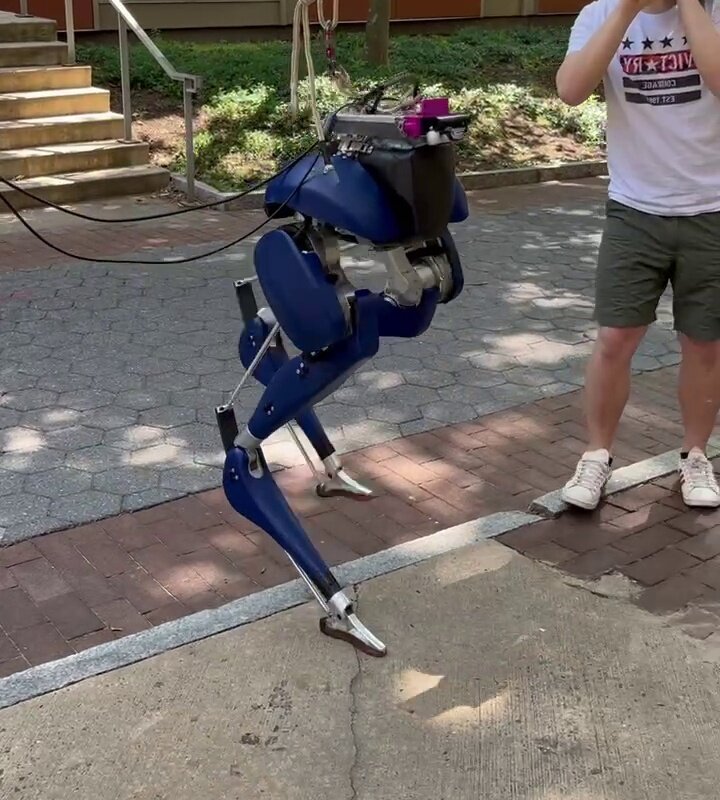}
	\includegraphics[width=0.1\textwidth]{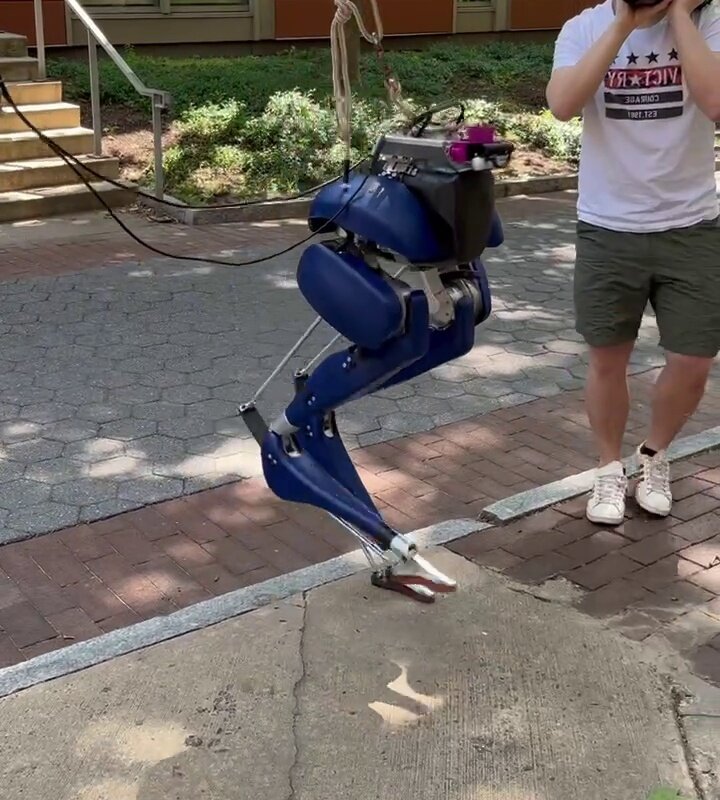}
	\includegraphics[width=0.1\textwidth]{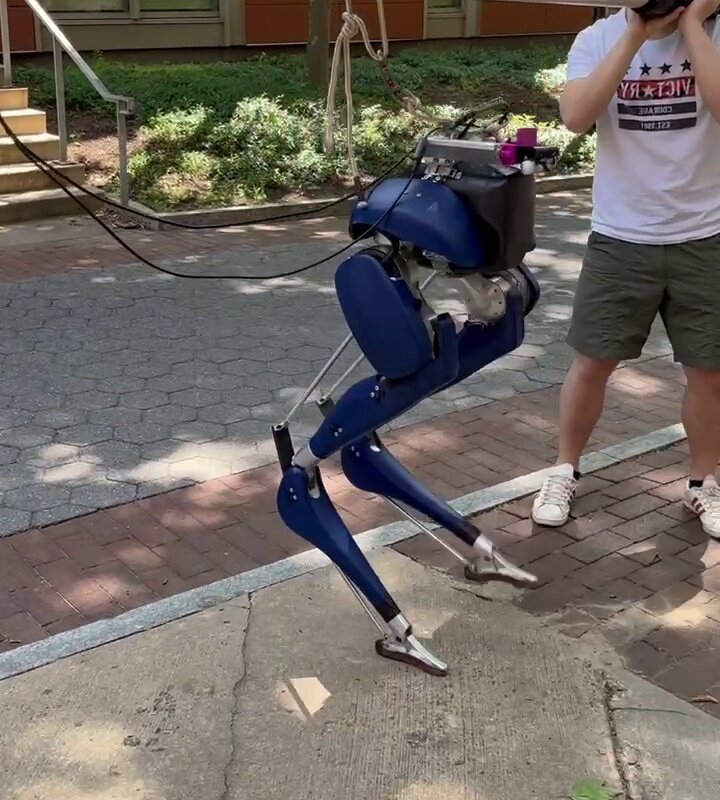}
	\includegraphics[width=0.1\textwidth]{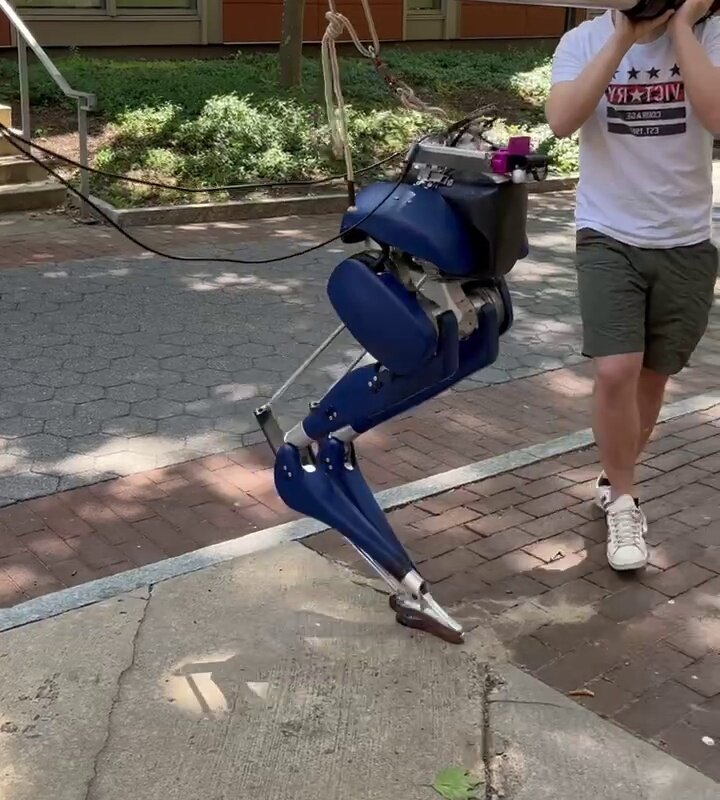}
	\includegraphics[width=0.1\textwidth]{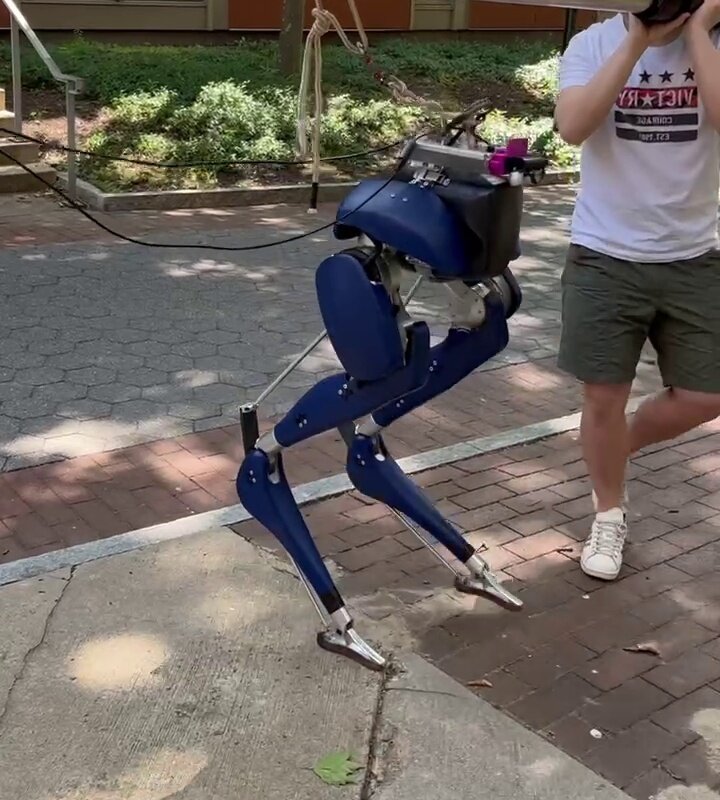}
	\includegraphics[width=0.1\textwidth]{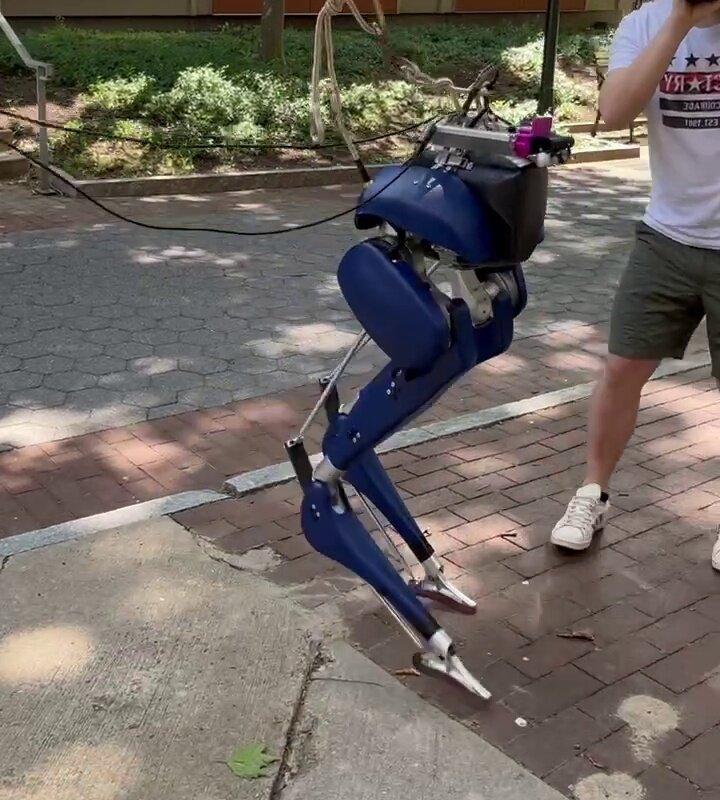}\\
	\includegraphics[width=0.1\textwidth]{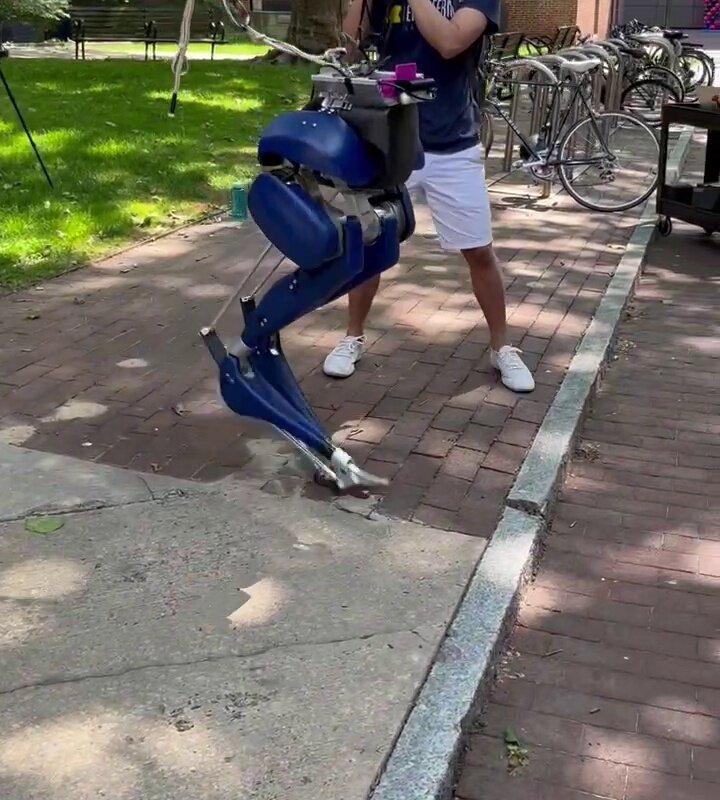}
	\includegraphics[width=0.1\textwidth]{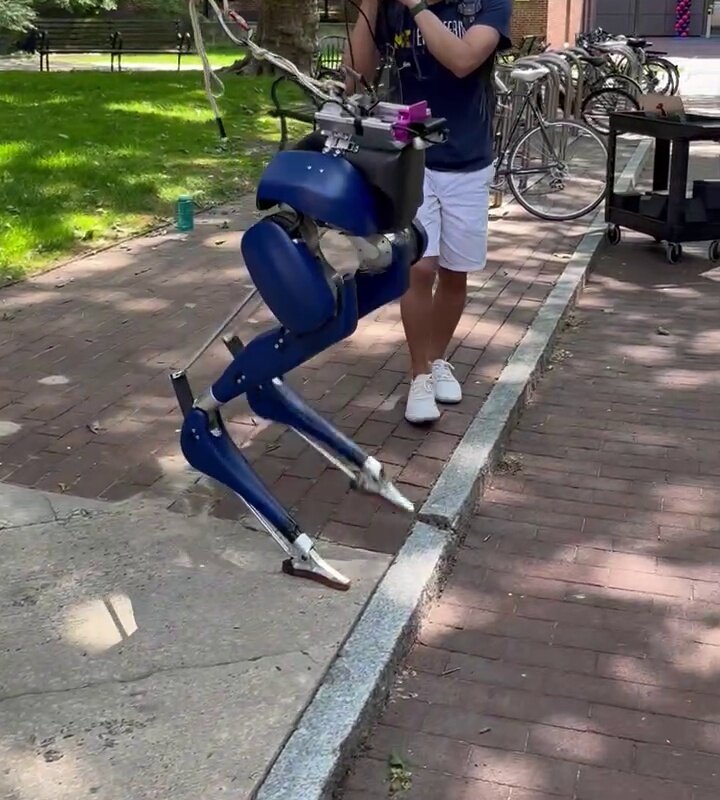}
	\includegraphics[width=0.1\textwidth]{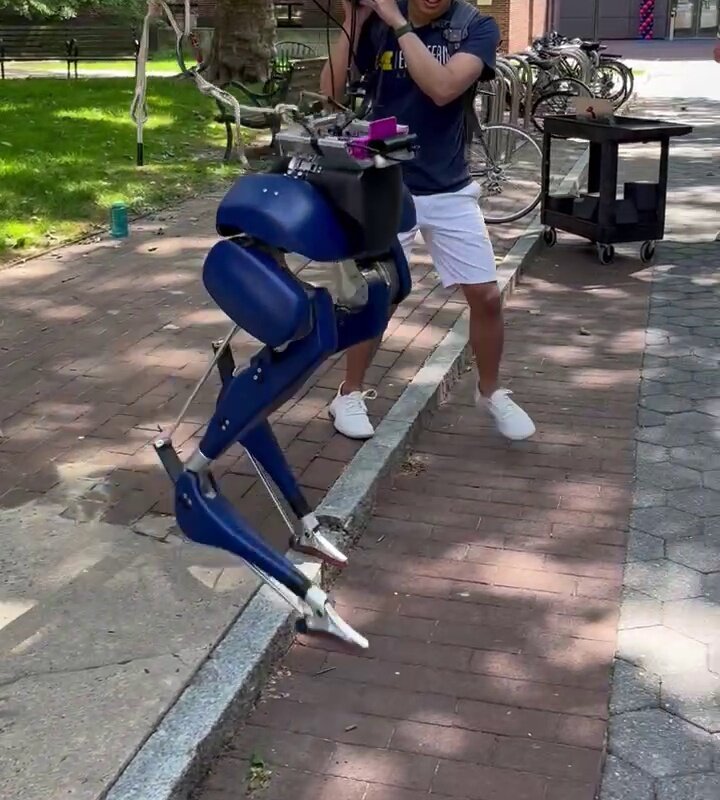}
	\includegraphics[width=0.1\textwidth]{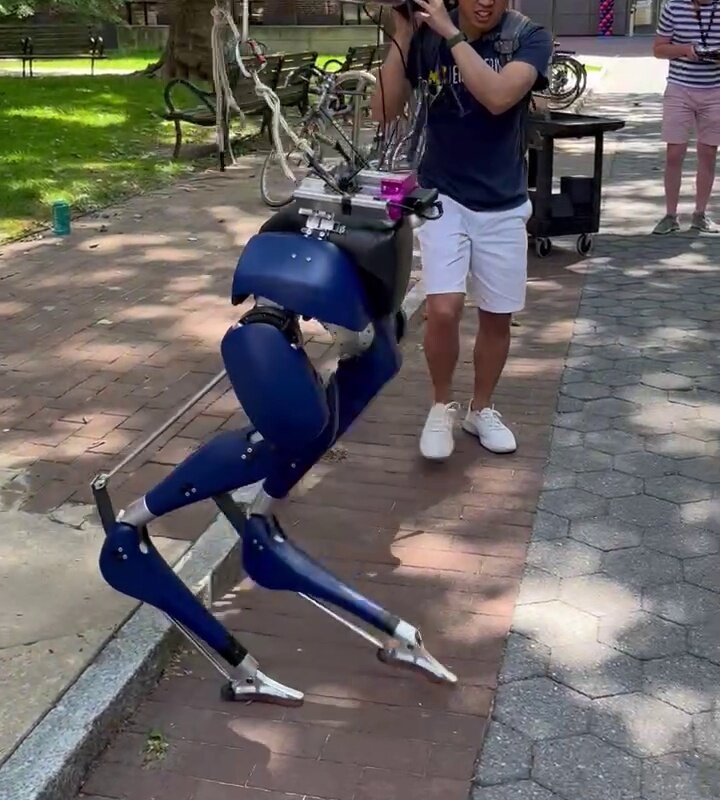}
	\includegraphics[width=0.1\textwidth]{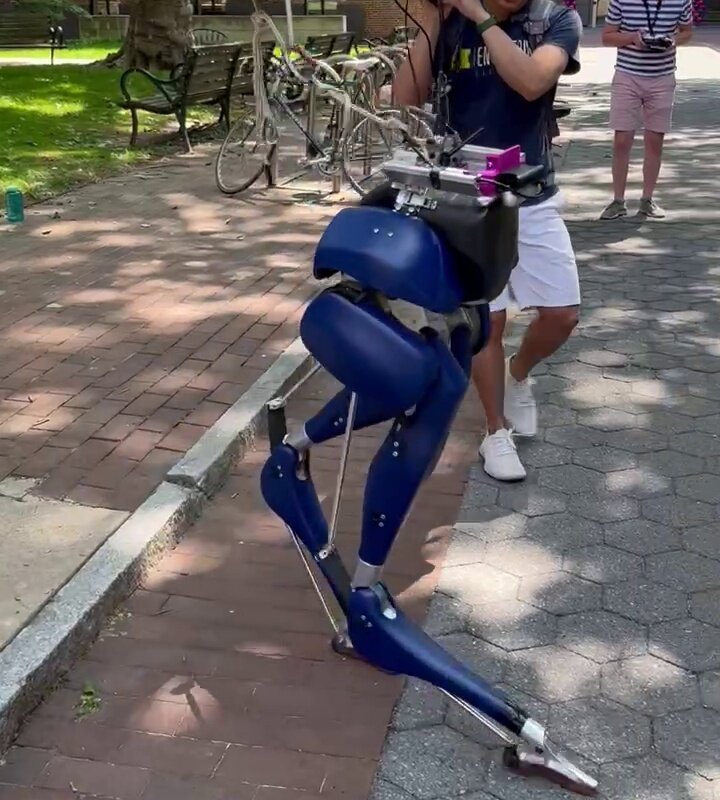}
	\includegraphics[width=0.1\textwidth]{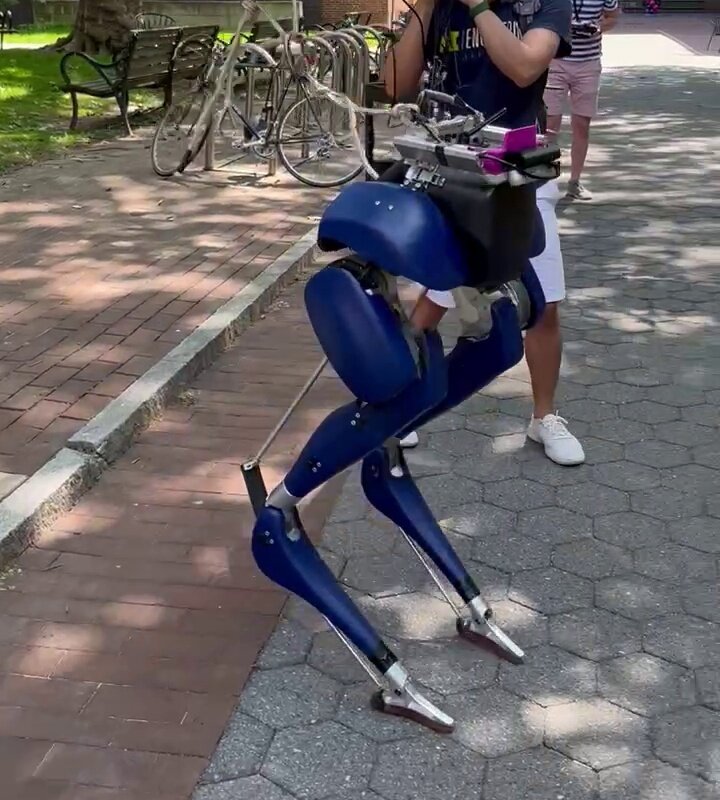}
	\includegraphics[width=0.1\textwidth]{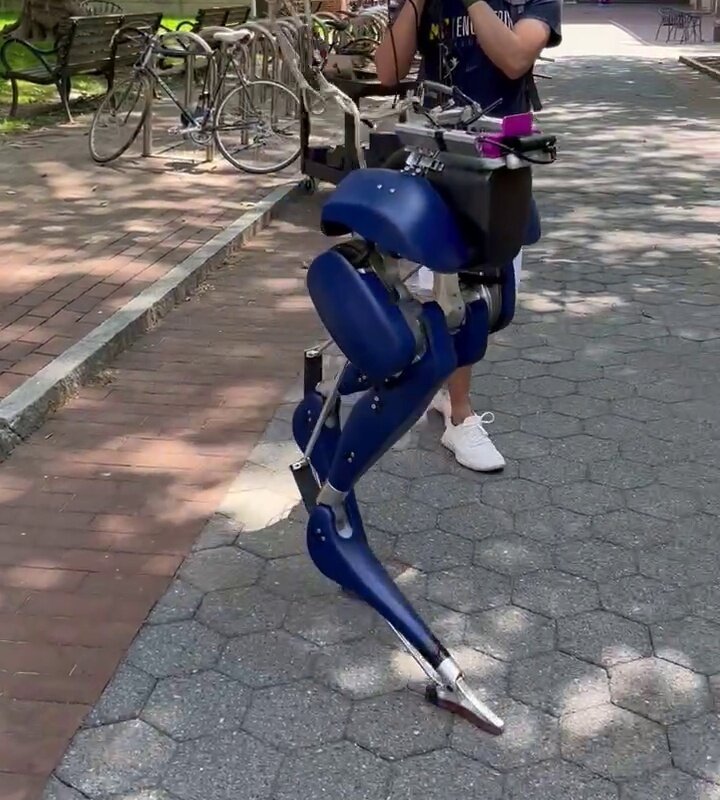}
	\includegraphics[width=0.1\textwidth]{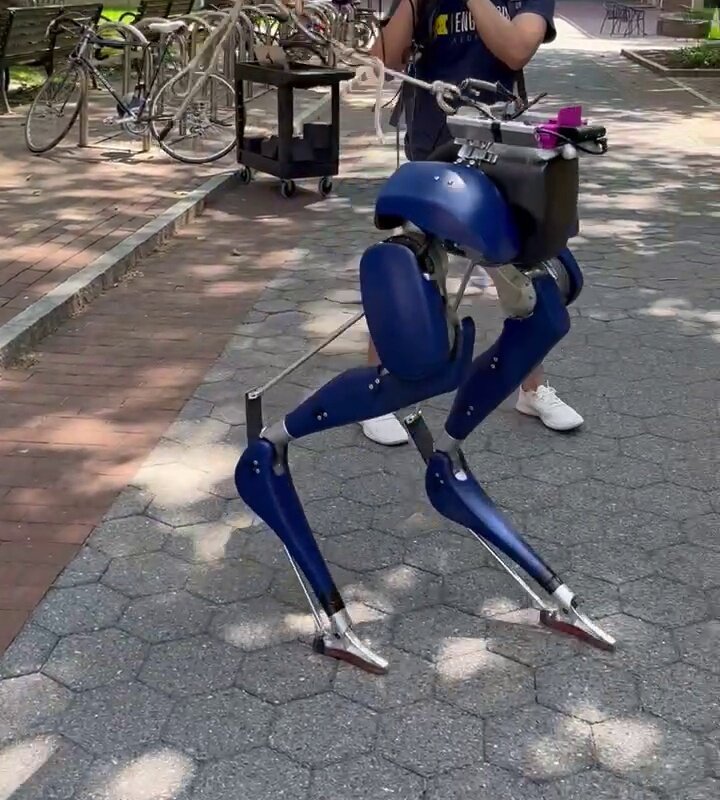}
	\includegraphics[width=0.1\textwidth]{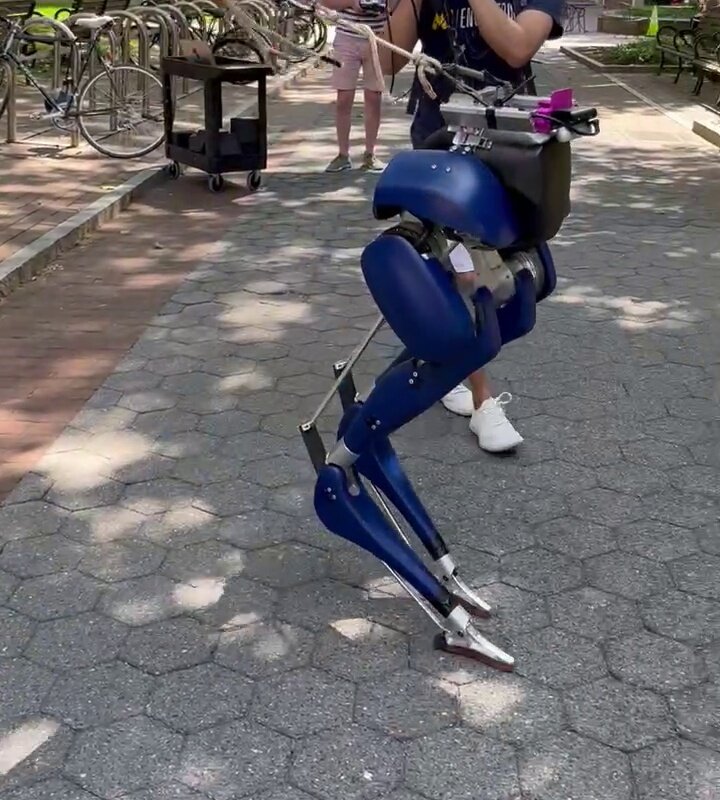}
	\caption{Motion tiles showing Cassie stepping up (top row) and then down (bottom row) a curb using the perception and control stack outlined in this paper. Neither the controller nor the perception stack have any prior knowledge of the environment.}
	\label{fig:motion_tiles}
\end{figure*}

We demonstrate our walking controller ascending and descending a 9 cm. curb with vision in the loop.  One trial can be seen in \cref{fig:motion_tiles}, and additional trials can be seen in the supplemental video \cite{AcostaBipedalYoutube}. While we achieved multiple successful trials, several interactions between the controller and perception stack make the system brittle in practice, as detailed in \cref{sec:challenge}.

\subsubsection{Controller Solve Times}
\begin{figure}[h]
	\centering
	\includegraphics[width=0.235\textwidth, valign=t]{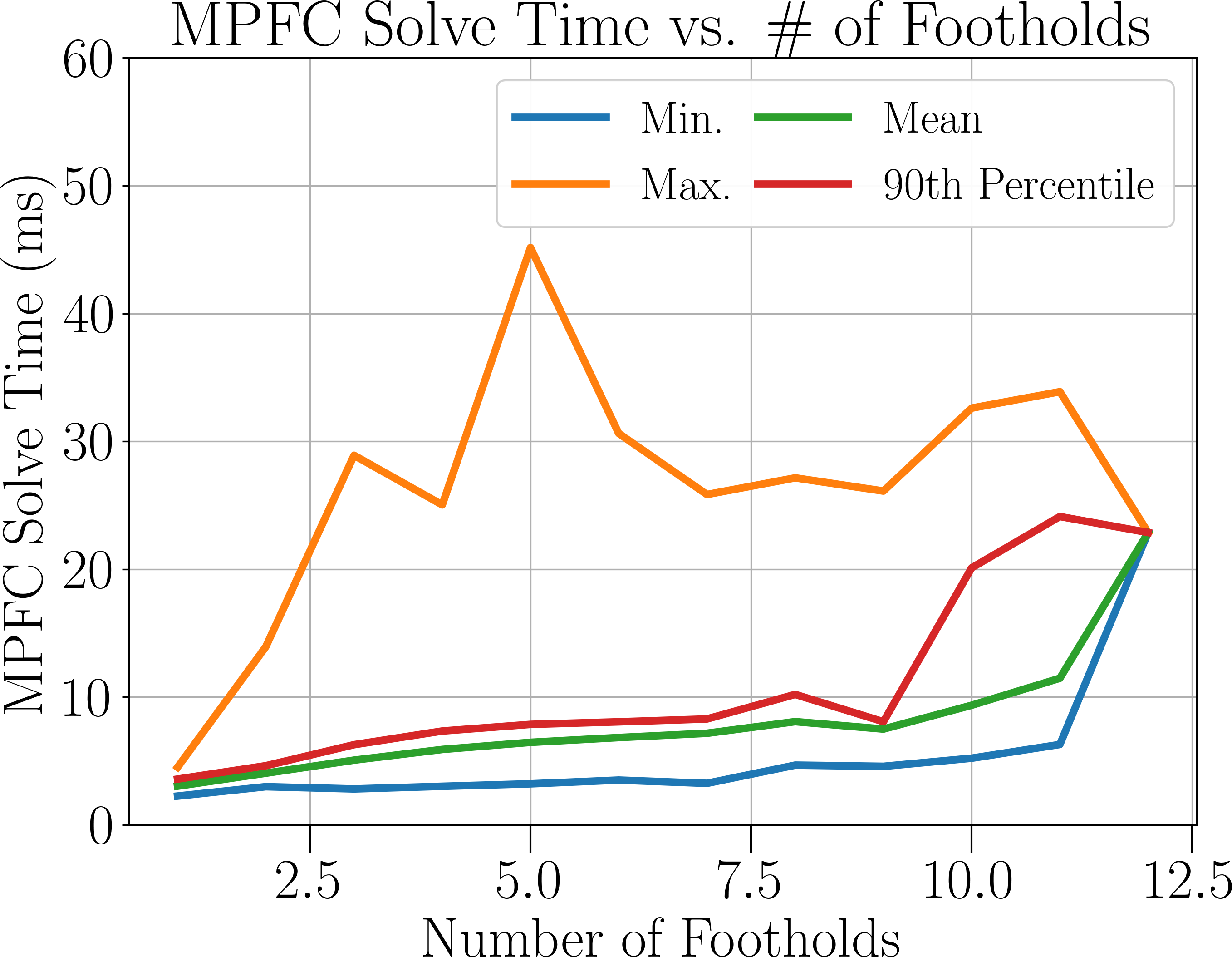}
    \includegraphics[width=0.235\textwidth, valign=t]{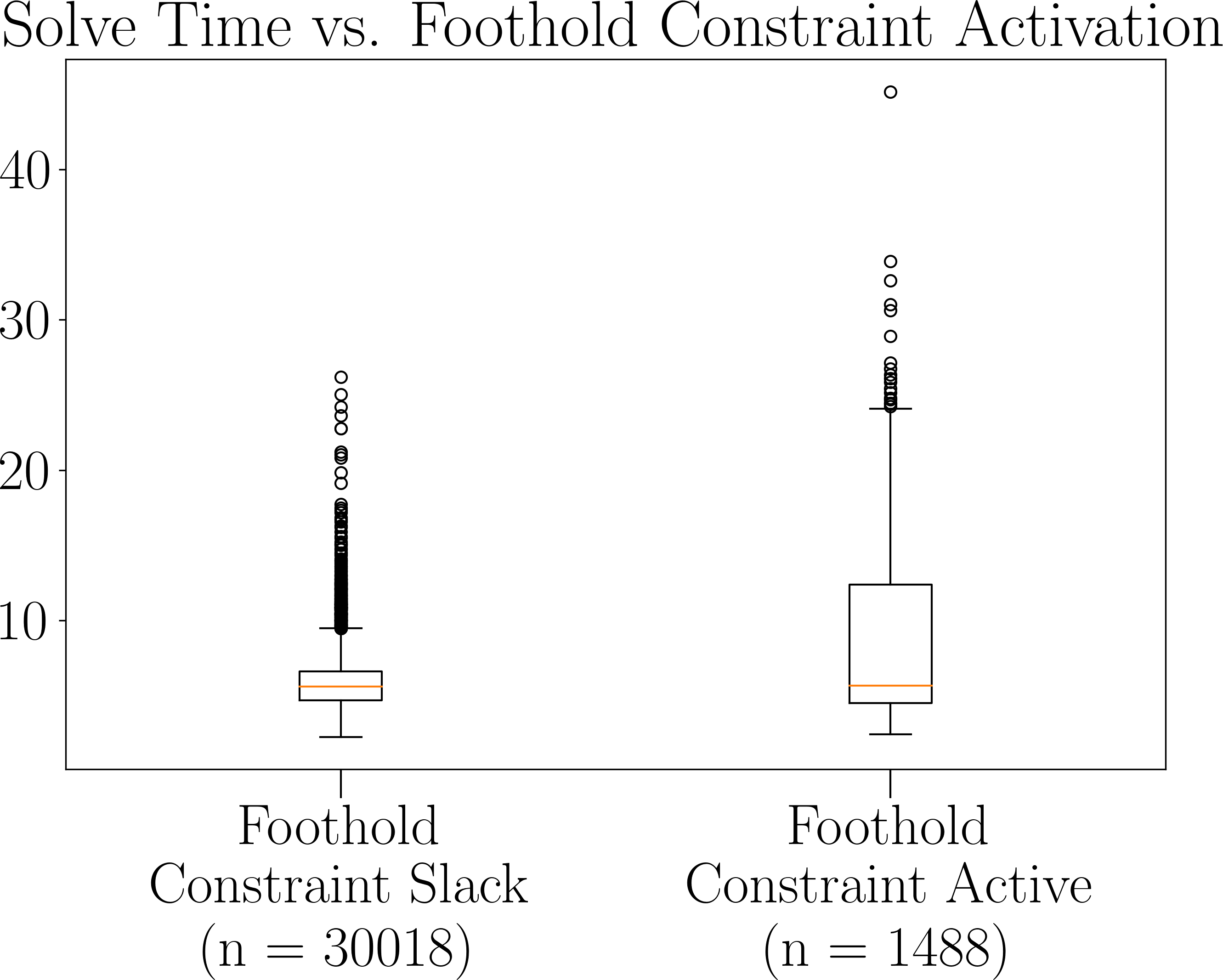}
	\caption{Analysis of MPFC solve times over 6 trials totaling 3:35 minutes of walking and 8 surface transitions. Left: Plot of solve time vs. the number of foothold constraints. Below 10 footholds, the average solve time increases slowly and approximately linearly. Right: Box and whisker plot showing the relationship between constraint activation and solve time. Solve time increases when the optimal footstep is on the boundary of a stepping stone, as the convex relaxation of the MIQP is no longer optimal\vspace{-0.4cm}} 
	\label{fig:solve_time}
\end{figure}

Since the computational complexity of the MPFC scales with the number of foothold constraints, and fast re-planning is required for stable walking, we analyze the effect of the number of potential footholds on solve time (\cref{fig:solve_time}). While the minimum, mean, and 90th percentile solve times are similar, and all increase slowly up to 9 footholds, the maximum solve time is 2.5-5 times higher. These occasional long solve times can be up to 10 percent of a single stance phase, and introduce torque spikes when tracking commanded foot position with high feedback gains. We also examine the relationship between foothold constraint activation and solve time. When the MPFC solution contains a footstep on a foothold boundary, the convex relaxation of the MIQP is no longer optimal, leading to longer solve times. This scenario makes up 4.9 percent of the data in \cref{fig:solve_time}, and will increase with more challenging terrains.

\section{Implementation Lessons}
\label{sec:challenge}
In this work, we synthesized a number of mature or maturing ideas. We combined LIP based footstep control and MIQP based kinematic footstep planning to design a novel walking controller. We informed this controller with a state of the art elevation mapping and convex planar decomposition framework. Here, we discuss successes and shortcomings of these methods and how they integrate into the full perceptive locomotion system. 

MPFC demonstrates yet again the effectiveness of (A)LIP based foot placement control for underactuated walking. When using MPFC on flat terrain, it is capable of achieving robust, dynamic walking similar to our baseline ALIP walking controller based on \cite{gongOneStepAheadPrediction2021}. Our simulation results also show that with ideal terrain information, incorporating mixed integer footstep constraints expands the capabilities of ALIP based walking control to discontinuous terrain. Fast solve times point to the fact that more advanced solvers are making it possible to use MIQPs for high-rate control tasks.

Where we experienced challenges was our controller's sensitivity to perception error and noise. The GPU based elevation mapping framework has so far been applied primarily to quadruped robots, which are lighter, more stable, and less susceptible to the effects of impacts when walking. Additionally, existing controllers with real-time perceptive footstep planning generally plan footholds about once per stride \cite{calvertFastAutonomousBipedal2022}, \cite{corberesPerceptiveLocomotionWholeBody2023}, \cite{grandiaPerceptiveLocomotionNonlinear2022}, making them less susceptible to noise in the perception system. The MPFC foothold constraints, on the other hand, are updated in real time, and are the direct output of a 30 Hz perception system. Therefore the whole system's performance is dependent on accurate, consistent perception. 
\subsection{Elevation Mapping Artifacts}
Errors and noise in the elevation map can propagate through the perception pipeline to the controller, leading to failure. When walking on grass, for example, the stance foot drift correction conflicts with the height of the point cloud, as Cassie's foot rests below the top of the grass. This leads to discontinuities in the height map when walking on flat grass. Additionally, the controller is sensitive to errors in the estimated ground height, which alter the swing foot touchdown time, ultimately causing the swing-foot to miss its commanded target.  

\subsection{Safety Margins}
Because we model Cassie's feet as points in the MPFC, we rely on a safety margin in the planar segmentation algorithm to account for the length of the blade foot. Tuning this margin is quite difficult in practice, as there is a trade off between collision risk and traversable area. Insufficient margin leads to collisions when stepping up onto a higher surface, but unnecessary margin leads to terrain gaps too large to cross at reasonable walking speeds. We experienced both of these failure modes with the same margin, depending on the commanded velocity and initial conditions of the step. While we attempted shaping the swing foot trajectory to avoid collisions, these trajectories were difficult to track due to large accelerations needed to start the swing phase moving away from the target position. The walking speed dependence of this effect increased the skill and concentration required from the operator. Future work will consider reformulating the MPFC cost function to address these challenges, as well as allowing for asymmetric safety margins in the planar polygon contour extraction process depending on the relative height of each surface. 

\subsection{Perception Noise}
When the planned footsteps are at or near the boundary of a foothold, noise in the robot state or foothold boundaries can cause sudden jumps in the impending footstep position by changing which foothold sequence is optimal. To avoid causing large jumps which would be impossible to track, we introduce an additional bounding box constraint on the next footstep position. The constraint is applied when 250ms remain in the swing phase, and constrains the upcoming footstep to a bounding box with half length 10 cm, centered at the footstep solution from the most recent MPFC solve.
	\section{Conclusions and Future Work}
We present a new model predictive footstep controller which allows underactuated bipeds to walk on constrained terrain without predefined foothold sequences. We formulate our controller as a single Mixed Integer Quadratic Program which can be solved online faster than the 30 Hz rate of the realtime perception system used to model the terrain as convex polygonal footholds. We demonstrate the controller on Cassie with a fully integrated vision system. 

Future work will focus on improving the reliability of the perception pipeline to give a consistent and accurate terrain representation while maintaining its real-time performance, and improving the robustness of the MPFC and OSC to perception error.

	\section{Acknowledgements}
	We thank Yu-Ming Chen and William Yang for assistance with experiments and years of collaboration and mentorship working on Cassie. We also thank Chandravan Kunjeti, Wei-Cheng Huang, and Alp Aydinoglu for help with experiments and helpful discussions. This material is based upon work supported by the National Science Foundation Graduate Research Fellowship Program under Grant No. DGE-1845298. Toyota Research Institute also provided funds to support this work.

	\bibliographystyle{IEEEtran} 
	\small
	\bibliography{references}

\end{document}